\documentclass[preprint]{article}

\PassOptionsToPackage{numbers,sort&compress}{natbib}

\usepackage{neurips_2025}

\usepackage[utf8]{inputenc} 
\usepackage[T1]{fontenc}    
\usepackage{hyperref}       
\usepackage{url}            
\usepackage{booktabs}       
\usepackage{amsfonts}       
\usepackage{nicefrac}       
\usepackage{microtype}      
\usepackage{xcolor}         
\usepackage{bbding}
\usepackage{pifont}
\usepackage{amsthm}
\usepackage{amsmath}
\usepackage{multirow}
\usepackage{graphicx}
\usepackage{enumitem}
\usepackage{subfigure}
\usepackage{algorithm}
\usepackage{algorithmic}
\usepackage{threeparttable}
\usepackage{lineno}
\nolinenumbers

\theoremstyle{definition}
\newtheorem{definition}{Definition}

\title{Satellites Reveal Mobility: A Commuting Origin-destination Flow Generator for Global Cities}

\author{%
  Can Rong\textsuperscript{1} \quad
  Xin Zhang\textsuperscript{1} \quad
  Yanxin Xi\textsuperscript{2} \quad
  Hongjie Sui\textsuperscript{1} \quad
  Jingtao Ding\textsuperscript{1} \quad
  Yong Li\textsuperscript{1} \\
  \textsuperscript{1}Dept. Electronic Engineering, BNRist, Tsinghua University \\
  \textsuperscript{2}Dept. Computer Science, University of Helsinki
}

\begin{document}

\maketitle

\begin{abstract}
  Commuting Origin-destination~(OD) flows, capturing daily population mobility of citizens, are vital for sustainable development across cities around the world. However, it is challenging to obtain the data due to the high cost of travel surveys and privacy concerns. Surprisingly, we find that satellite imagery, publicly available across the globe, contains rich urban semantic signals to support high-quality OD flow generation, with over 98\% expressiveness of traditional multisource hard-to-collect urban sociodemographic, economics, land use, and point of interest data. This inspires us to design a novel data generator, GlODGen, which can generate OD flow data for any cities of interest around the world. Specifically, GlODGen first leverages Vision-Language Geo-Foundation Models to extract urban semantic signals related to human mobility from satellite imagery. These features are then combined with population data to form region-level representations, which are used to generate OD flows via graph diffusion models. Extensive experiments on 4 continents and 6 representative cities show that GlODGen has great generalizability across diverse urban environments on different continents and can generate OD flow data for global cities highly consistent with real-world mobility data. We implement GlODGen as an automated tool, seamlessly integrating data acquisition and curation, urban semantic feature extraction, and OD flow generation together. It has been released at \url{https://github.com/tsinghua-fib-lab/generate-od-pubtools}.
  
\end{abstract}

\section{Introduction} \label{sec:Introduction}

Commuting origin-destination~(OD) flows profile the regular population movement in daily life between every two urban regions within a given urban area~\cite{rong2023interdisciplinary}, providing a critical foundation for various applications. For example, OD flows are essential inputs for simulations and analyses in traffic management and urban planning, serving as the travel demands of citizens for developing more informed policies~\cite{cats2014dynamic,sun2013understanding}. Recent studies have also explored the potential of OD flows in supporting research related to the United Nations Sustainable Development Goals~(SDGs) on community detection~\cite{chen2024characterizing}, urban resilience~\cite{ribeiro2019urban,haraguchi2022human}, pubic health~\cite{jia2020population} and environmental protection~\cite{zeng2024carbon}, further demonstrating their importance. Therefore, obtaining OD flows holds great significance for cities around the world, especially as they increasingly support interdisciplinary research efforts that span traditional and emerging urban challenges~\cite{rong2023interdisciplinary,barbosa2018human}.

However, it is very challenging to obtain this valuable data. Traditional methods for obtaining OD flows, such as door-to-door travel surveys~\cite{rong2023interdisciplinary} and the aggregation of large-scale individual mobility trajectories~\cite{calabrese2011estimating,iqbal2014development,gundlegaard2016travel,pan2006cellular}, are often infeasible in target cities due to high costs and privacy concerns. Although recent studies have explored physical~\cite{zipf1946p,simini2012universal,barbosa2018human} and computational~\cite{pourebrahim2019trip,robinson2018machine,simini2021deep,liu2020learning,rong2023goddag,rong2023city,ronglarge} models for OD flow generation, these approaches face key limitations. Physical models rely solely on population distribution and adopt overly simplistic assumptions, ignoring urban semantic differences and resulting in limited accuracy. Computational models perform better but still rely on hard-to-obtain, expensive, and fine-grained inputs such as sociodemographics, infrastructure data, land use, and points of interest~(POIs), which limits their applicability to only one or a small number of data-rich, developed cities.

\begin{figure}[t]
    \centering
    \includegraphics[width=\linewidth]{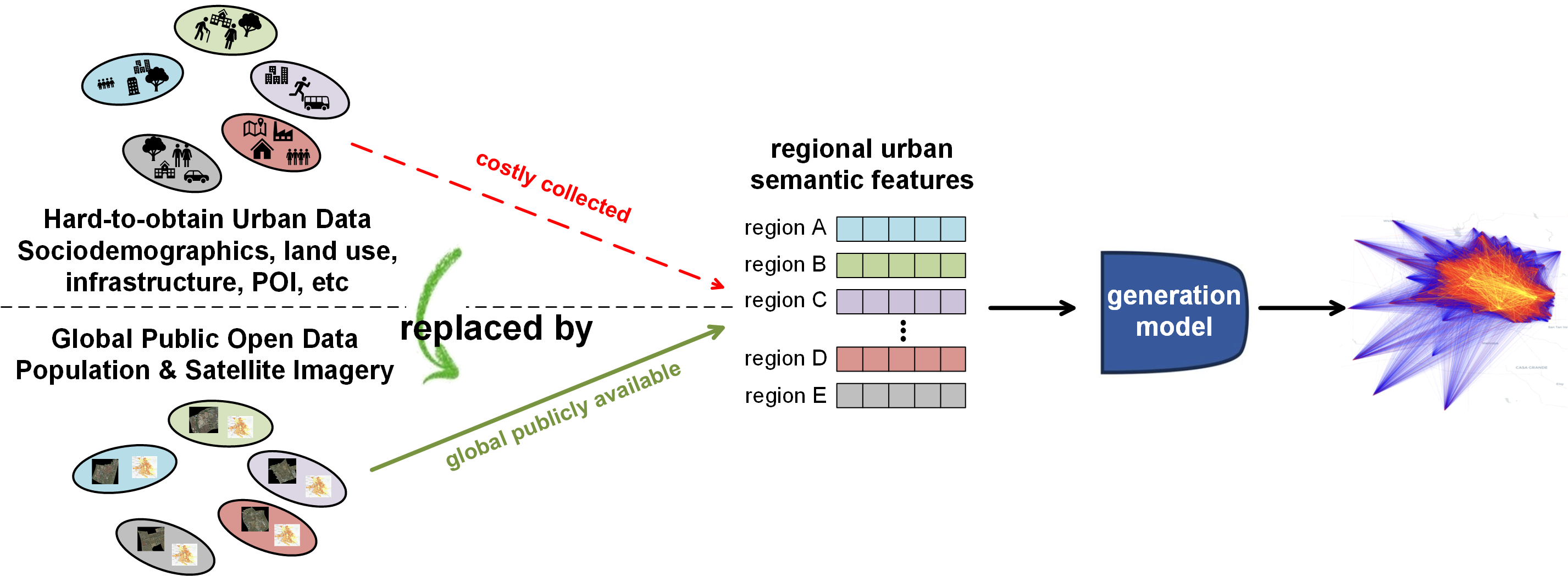}
    \caption{An illustration of replacing the traditional hard-to-collect urban data by easily accessible global public data~(e.g., satellite imagery and population data) in OD flow generation.}
    \label{fig:illustration}
\end{figure}

Recent studies~\cite{xi2023satellite,zhang2024uv,uzkent2019learning,han2020learning,jean2016combining,perez2017poverty,yeh2020using,he2018perceiving,xi2022beyond,han2020lightweight,hao2025urbanvlp,sharma2024kidsat,burke2021using,rolf2021generalizable,ahn2023human,abitbol2020interpretable} have demonstrated that satellite images, publicly available around the world, can provide important urban semantics related to human activities in urban regions. For example, residential structures identifiable in satellite imagery often indicate regions with high outbound commuting demand, while the presence of commercial buildings typically corresponds to regions with concentrated inbound flows. In contrast, sparsely built-up areas tend to exhibit lower levels of human activity and mobility. Therefore, satellite imagery holds the potential to serve as an alternative to the hard-to-obtain inputs required by traditional computational OD generation models, and supports the generation of OD flows for cities with limited data, as shown in Figure~\ref{fig:illustration}.

Accordingly, we propose GlODGen, which incorporates satellite imagery combined with population data as input for representing urban regions, and leverages recently developed vision-language geo-foundation models~\cite{zhou2024towards,li2024vision} to extract high-quality semantic features for OD flow generation in any city worldwide. Specifically, satellite imagery is first preprocessed by cropping and stitching it according to the boundaries of each urban region. This step removes irrelevant pixels and avoids the noise outside the regions. Then, with the help of RemoteCLIP~\cite{liu2024remoteclip}, the semantic features of urban regions are extracted. Finally, the extracted regional features are combined with population data to profile the urban space as input for the state-of-the-art OD flow generation model, WEDAN~\cite{ronglarge}, and generate OD flows. To investigate the validity of adopting only the publicly available satellite imagery and population data as input, we conduct experiments on representative urban areas in the U.S., Europe, China, Brazil, and Africa. We surprisingly find that totally publicly available data can serve as a perfect alternative to the hard-to-obtain inputs mentioned above and generate OD flows with high consistency with real-world data. This may shed light on the potential for extending OD flow generation to any city worldwide with the help of ubiquitous satellite imagery and population data, which is of great significance for the sustainable development of cities around the world. To facilitate practical use, we release GlODGen as an open-source tool at \url{https://github.com/tsinghua-fib-lab/generate-od-pubtools}, which seamlessly automates the collection and preprocessing of satellite imagery and population data, urban semantic feature extraction, and OD flow generation together.

The contributions of this paper can be summarized as follows:
\begin{itemize}[leftmargin=*]
    \item We find that publicly available satellite imagery and population data can replace hard-to-obtain urban data, such as sociodemographics, infrastructure data, land use, and POIs, to profile the urban space for inferring human mobility and generate OD flows.
    \item We propose GlODGen, a novel OD flow generator that leverages vision-language geo-foundation models to extract high-quality semantic features from satellite imagery and generate OD flows via graph diffusion models.
    \item We implement GlODGen as an efficient tool, which integrates data acquisition, curation and preprocessing, urban semantic feature extraction, and OD flow generation together. With only the boundaries of urban regions given, GlODGen can provide the OD flow data for cities of interest with \textit{only a single line of code}.
    \item We conduct experiments on representative urban areas around the world to demonstrate the effectiveness of GlODGen, which may pave the way for research on generating OD flows for global cities and contribute to lowering the barrier of data access for sustainable urban development.
\end{itemize}
\section{Preliminaries} \label{sec:Preliminaries}

\subsection{Definitions and Problem Formulation} \label{sec:Definitions}
\begin{definition}
    \textbf{City.} A city refers to a large, integrated urban area such as New York City, London, or Beijing. Our study focuses on commuting movements that occur within the spatial boundaries of a city.
\end{definition}

\begin{figure}[t]
    \centering
    \subfigure[]{
        \includegraphics[width=0.3\linewidth, height=0.23\linewidth]{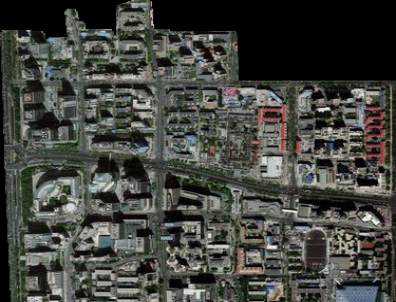}
    }
    \subfigure[]{
        \includegraphics[width=0.3\linewidth, height=0.23\linewidth]{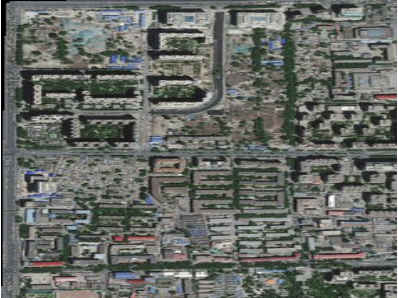}
    }
    \subfigure[]{
        \includegraphics[width=0.3\linewidth, height=0.23\linewidth]{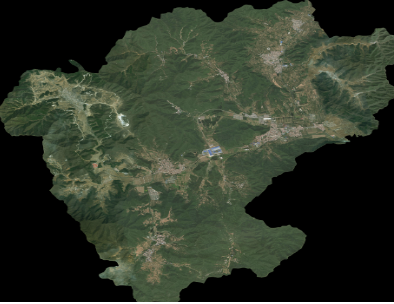}
    }
    \caption{Cases of satellite imagery for urban regions.}
    \label{fig:example}
\end{figure}

\begin{definition}
    \textbf{Urban Region.} Urban regions $\mathcal{R}=\{ r_i | i=1,2,...,N \}$ are the basic spatial units within a city for the OD flow generation task. Population movements are modeled between these regions.

    Every region corresponds to its unique satellite imagery, which consists of 2D, bird's-eye views captured from satellites covering the region, as illustrated in Figure~\ref{fig:example}.
\end{definition}
\begin{definition}
    \textbf{Commuting OD Flow.} This refers to the number of people $\mathcal{F}_{r_{org},r_{dst}}$ who live in one urban region $r_{org}$ and work in another urban region $r_{dst}$. It captures the regular, static daily movement from home to the workplace and remains relatively stable for a long time.
\end{definition}

\textsc{Problem 1.} \textit{\textbf{Commuting OD Flow Generation.}} Given a city and its region division, generate the commuting OD flows $\mathcal{F}=\{ \mathcal{F}_{r_{org},r_{dst}} | r_{org},r_{dst} \in \mathcal{R} \}$ that represent the daily commuting patterns of the city. The generated OD flows should be as consistent as possible with real-world human mobility patterns.

\subsection{Related Works} \label{sec:Relatedwork}

\textbf{Urban Representation Learning via Satellite Imagery.} Satellite imagery has been demonstrated to carry rich information about human activities, making it a feasible basis for profiling urban regions and generating OD flows. Specifically, satellite imagery offers a comprehensive overview of urban development from a macro perspective, while also enabling detailed analysis of regional contents at a micro level. At a coarse spatial level, satellite imagery has been leveraged to infer socio-economic indicators through tailored supervised and self-supervised learning methods, facilitating the extraction of high-level semantic representations of urban areas. These indicators include poverty levels~\cite{ayush2020generating,ayush2021efficient,han2020learning,jean2016combining,perez2017poverty,yeh2020using,xi2023satellite,hao2024urbanvlp,li2022predicting,zhang2024uv,han2020lightweight,sharma2024kidsat,burke2021using,ahn2023human,abitbol2020interpretable,xi2022beyond}, crop yields~\cite{m2019semantic,martinez2021fully,russwurm2020self,wang2018deep,yeh2021sustainbench,you2017deep}, land cover~\cite{hong2020graph,uzkent2019learning,li2024m3luc}, commercial activeness~\cite{he2018perceiving,li2023learning}, and environmental metrics~\cite{xi2022beyond,yan2024urbanclip,xiao2024refound,feng2024citybench,shao2021one,zhang2024uv,rolf2021generalizable,hao2024urbanvlp}. At a finer spatial scale, satellite imagery has been extensively utilized to monitor urban geospatial features, enabling the identification of various socio-physical entities such as streets, airplanes, vehicles, and recreational facilities like baseball fields~\cite{xi2024pixels,xia2018dota,li2020object,xi2023satellite,zhong2024urbancross,xiao2024refound,feng2024citybench,sun2021pbnet,xiao2024refound,chen2017practical}. These capabilities collectively highlight the strong potential of satellite imagery as a powerful and scalable modality for comprehensively representing urban regions, thereby enabling the generation of high-quality OD flow data around the world.

\textbf{OD Flow Generation.} OD flow obtaining, the task of obtaining OD flows for urban areas of interest lacking human mobility data, is an important yet traditionally costly and time-consuming process that relies on travel surveys~\cite{rong2023interdisciplinary,axhausen2002observing,iqbal2014development}. Researchers have explored alternative data sources related to individual trajectories, such as call detail records~(CDRs)~\cite{calabrese2011estimating,iqbal2014development} and cellular network accesss~(CNAs)~\cite{gundlegaard2016travel,pan2006cellular} to efficiently access OD flows. However, these methods face challenges regarding data accessibility and privacy concerns. To avoid these issues, researchers have introduced model-based methods for generating OD flows, which are often categorized into two distinct types. The first is physical models, e.g., the gravity model~\cite{zipf1946p} and radiation model~\cite{simini2012universal}, which model the population movement by mimicking physical laws, such as Newton's law of universal gravitation and the emission and absorption of the radiation process. These models are overly simplistic and are unable to effectively model the complexity inherent in human mobility. The second category is data-driven computational models that leverage machine learning and deep learning techniques to generate OD flows based on various urban factors, including sociodemographics, POI distributions, etc.~\cite{pourebrahim2019trip,robinson2018machine,simini2021deep,liu2020learning,rong2023goddag,rong2023city}. These methods achieve superior performance via sophisticated model structures but the high requirements for input data limits their applicability, especially in underdeveloped areas. Our framework leverages global publicly available data to replace the hard-to-collect features for profiling urban regions and generating OD flows. This breaks the barrier of data accessibility while reserving the superior performance advantages of computational models.
\section{GlODGen: Commuting Origin-destination Flow Generator} \label{sec:method}
\begin{figure}[t]
    \centering
    \includegraphics[width=\linewidth]{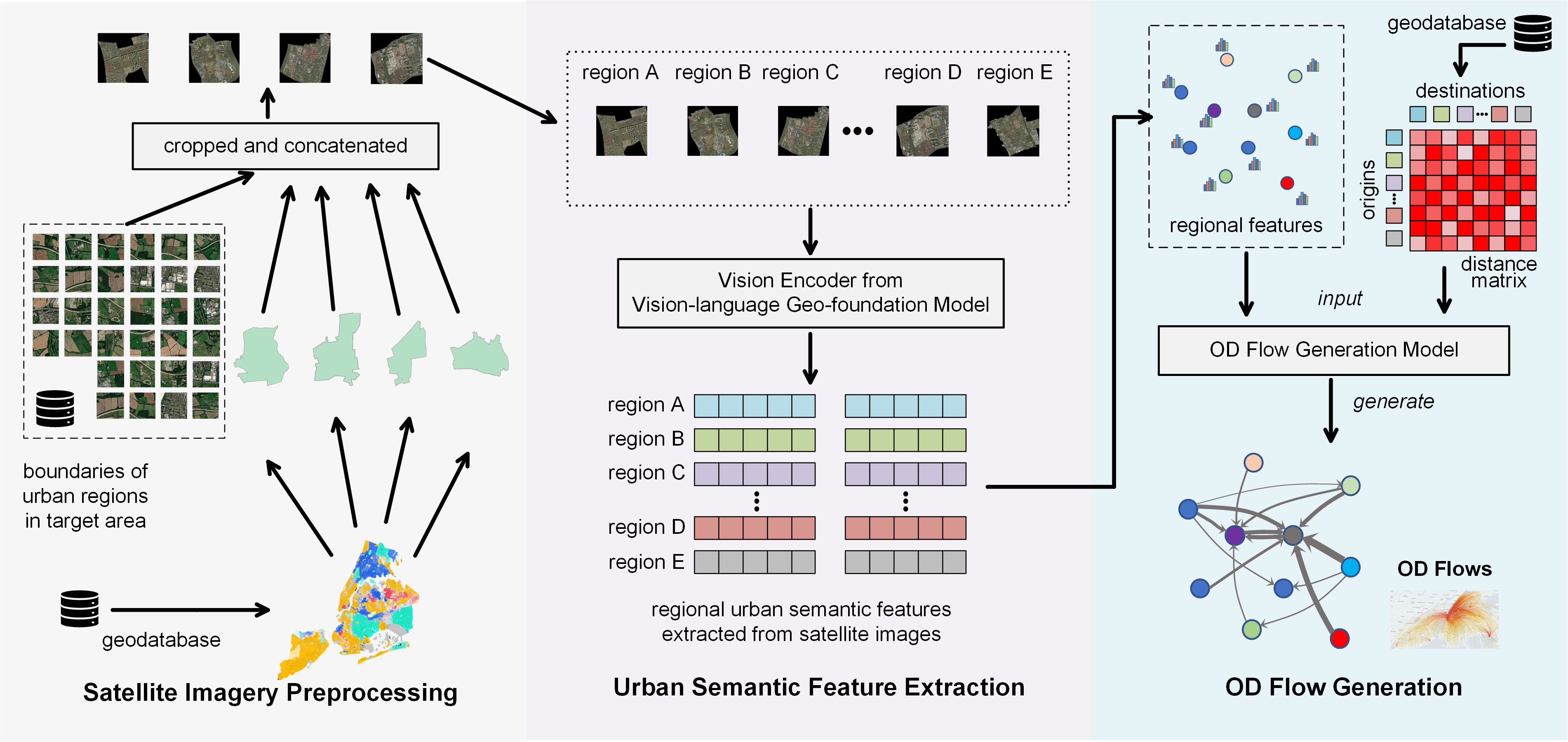}
    \caption{The framework and data pipeline of GlODGen.}
    \label{fig:systemoverview}
\end{figure}

In Figure~\ref{fig:systemoverview}, we illustrate the conceptual structure of the framework. It comprises three stages: \textit{satellite imagery preprocessing}, \textit{urban semantic feature extraction}, and \textit{origin-destination flow generation}. 

\textbf{Satellite Imagery Preprocessing.} For each region in the city, GlODGen first identifies specific tiles of satellite imagery corresponding to each region and downloads them from any online platform, such as Google Earth and Esri World Imagery. A tile refers to a smaller, square-shaped portion of a larger satellite image, which is created by dividing the satellite image of the whole world into manageable pieces. This tiling process facilitates more efficient obtaining and processing, especially when working with large-scale satellite images. With all tiles obtained, the next phase involves concatenation and cropping, shaping the images to match the exact contours of each region. This procedure is critical for ensuring that the imagery faithfully represents the geographical and physical attributes of each region, creating an accurate visual portrayal. During cropping, the image content beyond region boundaries is removed and zero-padded to preserve spatial accuracy and exclude irrelevant information, as formalized below:
\begin{equation}
    M_{r}(x, y) =
    \begin{cases}
    1, & \text{if } (x, y) \in \text{Boundary}(r) \\
    0, & \text{otherwise}
    \end{cases}
\end{equation}
where $(x,y)$ denotes the pixel coordinates, $M_{r}$ is the mask of region $r$. The image is then masked by $M_{r}$ to remove the irrelevant parts, which is detailed in the equation below:
\begin{equation}
    I_{\text{final}} = I_{\text{raw}} \odot M_{r}
\end{equation}
where $I_{\text{raw}}$ is the raw satellite image, $I_{\text{final}}$ is the final satellite image of region $r$, and $\odot$ denotes the element-wise multiplication operation. Once processed, the satellite images for the regions appear as depicted in Figure~\ref{fig:example}.

\textbf{Urban Semantic Feature Extraction.} After completing the previous step, every region in the specified area is assigned a satellite image to represent its unique characteristics and functions, as illustrated in Figure~\ref{fig:systemoverview}. However, the raw pixel data from the images are unsuitable for direct use as urban regional features, due to the presence of excessive noise and irrelevant, redundant information unrelated to human mobility. As such, a feature extraction model is needed to process the satellite imagery, enabling the distillation of the most significant semantic information related to OD flows. This model could be any structure, such as VGG~\cite{simonyan2014very}, ResNet~\cite{he2016deep}, or ViT~\cite{dosovitskiy2020image}, and we choose the vision encoder from the large multimodal model, which also has the special design to handle the satellite images, RemoteCLIP~\cite{liu2024remoteclip}. Pre-trained with natural language supervision for the vision encoder, the large multimodal model exhibits zero-shot capabilities, allowing it to extract versatile semantic features from satellite imagery for various downstream tasks. We adopt RemoteCLIP as the semantic feature extractor in our framework. Specifically, we input the preprocessed satellite image of each region to the vision encoder of RemoteCLIP, whose output is a 1024-dimension high-level feature that captures the urban semantics of that urban region. During this stage, we only use the vision encoder of RemoteCLIP with the pre-trained weights frozen. Because recent works have demonstrated that finetuning the foundation model on the specific task will lead to forgetting the general knowledge learned from the pre-training and result in overfitting~\cite{ding2022don,jatavallabhula2023conceptfusion,hong20233d}. This process is formalized as follows:
\begin{equation}
    E_r = \mathcal{V}(I_r),
\end{equation}
where $I_r$ is the satellite image of region $r$, $\mathcal{V}$ is the vision encoder of RemoteCLIP, and $E_r$ is the 1024-dimension feature vector of region $r$. The final input to the OD flow generation model is built by merging the semantic features, extracted from satellite images, with the population of region $r$, as shown below:
\begin{equation}
    X_r = [E_r, P_r],
\end{equation}
where $P_r$ means the population of $r$ and $[\dot]$ denotes a concatenation operation. $X_r$ is the final features profiling region $r$, which will be input into the OD flow generation model before being processed by a multi-layer perceptron~(MLP) whose parameters are trained while training the following OD flow generation model. In this way, the vision encoder can provide a general and versatile representation of the region, while the representation is further refined to fit the specific task of OD flow generation.

\textbf{Origin-destination Flow Generation.} We adopt WEDAN~\cite{ronglarge}, a state-of-the-art OD flow generation model based on graph denoising diffusion~\cite{jo2022score,niu2020permutation,vignac2022digress,haefeli2022diffusion,bergmeister2023efficient,fu2024hyperbolic}, to generate OD flows. In WEDAN, urban regions are represented as nodes and OD flows as weighted directed edges in a graph. The model follows a conditional generation paradigm, where node attributes serve as guidance for the denoising process and generating the directed edges and corresponding weights. In other words, the semantic characteristics of urban space are leveraged as conditioning inputs, guiding the denoising process to produce realistic and spatially consistent OD flows. Benefiting from its formulation of the city-wide OD network as a directed, weighted graph, WEDAN is capable of generating OD flows that closely align with real-world population mobility patterns. In our framework, the urban semantic feature $X_r$ extracted for each region is used as the node attribute input, guiding the generation of edge directions and weights during the denoising process. The procedure can be described as follows:
\begin{equation}\label{eq:ODdenoise}
    \begin{split}
    \begin{aligned}
    & p_{\theta} (\mathbf{F}^{t-1}|\mathbf{F}^t,\mathcal{C_\mathcal{R}}) = \mathcal{N} (\mathbf{F}^{t-1} ; \mu_{\theta}(\mathbf{F}^t,t,\mathcal{C_\mathcal{R}}) ,(1-\bar{\alpha}^t)\mathbf{I}),\\
    \end{aligned}
    \end{split}
\end{equation}
where
\begin{equation}\label{eq:miu_previous}
    \begin{split}
    \begin{aligned}
    & \mu_{\theta}(\mathbf{F}^t,t,\mathcal{C_\mathcal{R}}) = \frac{1}{ \sqrt{\alpha_t} } (\mathbf{F}^t - \frac{\beta_t}{\sqrt{1-\bar{\alpha}_t}} \epsilon_\theta(\mathbf{F}^t,t,\mathcal{C_\mathcal{R}}) ) .\\
    \end{aligned}
    \end{split}
\end{equation}
In the formulas, $\mathcal{C_\mathcal{R}} = \{ X_r | r \in \mathcal{R} \}$ is the semantic features of urban regions within the city, $t$ is the diffusion step, $\mathbf{I}$ is the identity matrix, $\epsilon_\theta$ means denoising networks, $\mu_{\theta}$ denotes $\mu$ of the Gaussian distribution, $ \alpha_t = 1-\beta_t$ and $\bar{\alpha}_t = \prod_{i=1}^t \alpha_i$ refer to the noise scheduler. The denoising network utilized here is the graph transformer network~\cite{dwivedi2020generalization}, which predicts the noise needed to be removed from the current graph state $\mathbf{F}^t$ to reach the previous state $\mathbf{F}^{t-1}$.

\section{Experiments} \label{sec:experiments}

In this section, we conduct experiments to answer two key research questions:
\begin{itemize}[leftmargin=*]
    \item {\textbf{RQ1:}} For precision, can entirely public data, satellite imagery, and population data, sufficiently represent urban spatial characteristics and support the generation of high-quality OD flows?
    \item {\textbf{RQ2:}} For generalizability, does GlODGen demonstrate cross-continental transferability, enabling it to generate OD flows in diverse global cities with the help of global public input data?
\end{itemize}

All experiments were conducted on a single NVIDIA GeForce RTX 4090 GPU (24GB) and an Intel Xeon Platinum 8358 CPU @ 2.60GHz. For all trainable models, we performed the grid search to select optimal hyperparameters. The same training, validation, and testing splits were used when training all models and evaluating the performance.

\subsection{Performance of Public Satellite Imagery and Population Data~(RQ1)} \label{sec:rq1}

We begin by evaluating the effectiveness of publicly available satellite imagery and population data in supporting OD flow generation. Specifically, we compare the performance of several existing OD flow generation models using two different types of input: (1) traditional fine-grained features that are often costly or difficult to obtain, and (2) publicly accessible features. In this experiment, we use detailed sociodemographic attributes, economic indicators, and POI distributions as the traditional inputs, and satellite imagery combined with population data as the public inputs. To evaluate representational capacity across input types, we apply two groups of existing models commonly used for OD flow generation: physical models, including the classical gravity and radiation models, which rely solely on population distribution act as lower-bound reference of performance; Computational models, including Random Forest~\cite{pourebrahim2019trip}, DeepGravity~\cite{simini2021deep}, GMEL~\cite{liu2020learning}, NetGAN~\cite{bojchevski2018netgan}, and WEDAN~\cite{ronglarge}, which support richer semantic inputs and allow us to assess the performance impact of different input data sources. A detailed introduction of the models is provided in Appendix~\ref{apdx:modelintro}.

\textbf{Dataset.} In this part, we use the dataset~\cite{ronglarge} from the United States to conduct experiments. This dataset totally consists of 3,333 urban areas, including 3,233 counties and 100 metropolitans. Each urban area is associated with its region division: census tracts within counties and census block groups within metropolitans. The dataset contains two parts: (1) city characteristics in urban regions, including population structure, education level, poverty, income, vehicle ownership, and other socioeconomic indicators from the American Community Survey~(ACS) and (2) OD flow data between urban regions provided by the National Census Bureau through the Longitudinal Employer-Household Dynamics Origin-Destination Employment Statistics~(LODES)~\cite{uscensusbureau2024}.

\textbf{Experimental Settings.} To systematically evaluate the performance of different models and input types, we adopt root mean square error~(RMSE), normalized RMSE~(NRMSE), and common part of commuting~(CPC) as metrics. We split the 3,333 urban areas into 8:1:1 proportions for training, validation, and testing. The evaluation metrics are averaged over all test urban areas. All experiments are repeated 5 times with different random seeds with average results and standard deviations reported.

\begin{table}[ht]
    \caption{Performance of existing OD flow generation models on test urban areas in the United States with different input data. Traditional hard-to-obtain urban data and global public data are put together for better comparison. The former's results are shown in the upper line, while the latter's are shown in the lower line.}
    \label{tab:dataperformance}
    \centering
    \begin{tabular}{@{}c|cc|cc|c@{}}
    \toprule
    \multicolumn{1}{l|}{}      & CPC$\uparrow$   & Perc.  & RMSE$\downarrow$ & Gap & NRMSE$\downarrow$ \\ \midrule \midrule
    Gravity Model                         & 0.321 $\pm$ 0.02 & -      & 174.0 $\pm$ 10.4& - & 2.222 \\
    Radiation Model                         & 0.347 $\pm$ 0.04 & -      & 196.9 $\pm$ 11.7 & - & 2.502 \\ \midrule
    \multirow{2}{*}{Random Forest}        & 0.494 $\pm$ 0.02& -      & 100.4 $\pm$ 6.5 & - & 1.282 \\
                               & \textbf{0.480} $\pm$ 0.03 & 97.1\% & 114.0 $\pm$ 7.1 & -13.5\% & 1.455 \\ \midrule
    \multirow{2}{*}{DeepGravity}        & 0.449 $\pm$ 0.01 & -      & 92.9 $\pm$ 9.9 & - & 1.186 \\
                               & \textbf{0.427} $\pm$ 0.05 & 95.1\% & 99.9 $\pm$ 12.0 & -7.5\% & 1.275 \\ \midrule
    \multirow{2}{*}{GMEL}      & 0.462 $\pm$ 0.01 & -      & 94.3 $\pm$ 4.0 & - & 1.204 \\
                               & \textbf{0.451} $\pm$ 0.02 & 97.6\% & 105.4 $\pm$ 7.9 & -11.8\% & 1.345 \\ \midrule
    \multirow{2}{*}{NetGAN}    & 0.517 $\pm$ 0.05 & -      & 89.1 $\pm$ 17.0 & - & 1.138 \\
                               & \textbf{0.468} $\pm$ 0.06 & 90.5\% & 98.0 $\pm$ 12.3 & -10.0\% & 1.251 \\ \midrule
    \multirow{2}{*}{WEDAN} & 0.634 $\pm$ 0.01 & -      & 64.06 $\pm$ 3.3 & - & 0.818 \\
                               & \textbf{0.623} $\pm$ 0.02 & 98.3\% & 67.88 $\pm$ 6.1 & -5.9\% & 0.867 \\ \bottomrule
    \end{tabular}
\end{table}

\textbf{Experimental Results.} As shown in Table~\ref{tab:dataperformance}, public data demonstrates performance close to that of traditional hard-to-obtain urban data. We can see that the performances of global public data are slightly less accurate than those of traditional hard-to-collect information, but the deviations are almost indistinguishable. Across all models based on the data-driven schema, public data achieve over 90\% of the performance of traditional features. This indicates that such data can further effectively profile and represent spatial characteristics in a manner comparable to traditional hard-to-obtain inputs. Moreover, computational models based on public data significantly outperform classic physical models, demonstrating the superior performance and widespread applicability of public data, including satellite imagery and population data.

\subsection{Generalization Across Continents~(RQ2)} \label{sec:rq2}

In this part, we conduct experiments that transfer the OD flow generation model from one trained continent to another and investigate the generalization ability of GlODGen with global publicly available satellite imagery and population data. Specifically, we use the data from the United States to train the GlODGen model, and then transfer it to the United Kingdom for evaluation. 

\textbf{Dataset.} The training data in the United States is the same as the dataset used in Section~\ref{sec:rq1}. The test data in the United Kingdom is from the Office for National Statistics~(ONS)~\cite{ONS2023}, which provides population-level commuting flows between all census units. Specifically, the data contains OD flows of 326 Local Authority Districts~(LAD). In each district, the whole area is divided into urban regions by the boundaries of Middle layer Super Output Areas~(MSOA). 

\textbf{Experimental Settings.}  We use 90\% urban areas in the United States as the training data and the remaining 10\% as the validation data to tune the hyperparameters. The trained GlODGen model is then transferred to the United Kingdom for generating OD flows. Generated OD flows will be compared with the ones from census data and evaluated by metrics introduced in Section~\ref{sec:rq1}. Metrics are the average of all LADs. We also introduce existing models to perform the same task and evaluate their performance as baselines. It is worth noting that existing models are only provided with population data and inter-region distances as input. This limitation is unavoidable, as fine-grained urban features that align across training and testing cities are typically unavailable in cross-continental experimental settings.

\textbf{Experimental Results.} As shown in Table~\ref{tab:US2UK}, the proposed framework can be transferred between different continents with a robust performance. Compared with the baseline models, our framework can achieve a promising performance and outperform baselines on all metrics with a large margin, i.e., 34.0\% improvement on CPC, and 28.5\% improvement on RMSE and NRMSE. This indicates that 1) public data, i.e., satellite imagery and population data, can effectively be transferred between different continents; 2) the proposed framework can improve the transferability of the computational OD flow generation models; 3) computational models based on public data can beat physical models in cross-continental transfer experiments, supporting a great improvement on global scale OD flow generation. This strongly supports the conclusion that our framework is highly suitable for generating OD flows across diverse urban areas worldwide.

\begin{table}[t]
    \caption{Performance comparison of baseline methods with geographically distributed population and our proposed framework profiling urban regions based on global public data on the United Kingdom. All the methods are trained in the United States and evaluated in the United Kingdom.}
    \label{tab:US2UK}
    \centering
    \resizebox{\textwidth}{!}{
    \begin{tabular}{c|cc|cc|cc}
    \toprule
    \textbf{Model} & \textbf{CPC$\uparrow$} & IMP. & \textbf{RMSE$\downarrow$} & IMP. & \textbf{NRMSE$\downarrow$} & IMP. \\
    \midrule\midrule
    Gravity Model          & 0.240 $\pm$ 0.01 & -27.5\% & 101.6 $\pm$ 3.42 & +48.9\% & 1.752 $\pm$ 0.06 & +48.9\% \\
    Radiation Model        & 0.323 $\pm$ 0.07 & -2.4\% & 211.6 $\pm$ 4.27 & -6.4\% & 3.647 $\pm$ 0.07 & -6.4\% \\
    \midrule
    Random Forest          & 0.334 $\pm$ 0.03 & +0.9\% & 223.2 $\pm$ 3.42 & -12.2\% & 3.847 $\pm$ 0.06 & -12.2\% \\
    DeepGravity           & 0.359 $\pm$ 0.04 & +8.5\% & 157.0 $\pm$ 7.90 & +21.1\% & 2.706 $\pm$ 0.13 & +21.1\% \\
    GMEL                  & 0.362 $\pm$ 0.01 & +9.4\% & 149.1 $\pm$ 10.00 & +25.0\% & 2.570 $\pm$ 0.17 & +25.1\% \\
    \midrule
    NetGAN                & 0.331 $\pm$ 0.06 & - & 198.9 $\pm$ 15.21 & - & 3.429 $\pm$ 0.26 & - \\ 
    GlODGen             & \textbf{0.485} $\pm$ 0.03 & +46.5\% & \textbf{72.68} $\pm$ 2.13 & +63.5\% & \textbf{1.253} $\pm$ 0.04 & +63.5\% \\
    \bottomrule
    \end{tabular}
    }
\end{table}

\subsection{Case Studies on Typical Urban Areas Worldwide~(RQ1 \& RQ2)}

To further validate the effectiveness of the proposed data generator, we collect human mobility-related data from all over the world and apply the OD flow generation to the corresponding urban areas. Specifically, we collect the data from the following typical urban areas and detailed data processing is provided in Appendix~\ref{apdx:datadoc}.
\begin{itemize}[leftmargin=*]
    \item \textbf{Europe.} We select London and Paris as representative cities in Europe. The OD flow data for London is sourced from ONS~\cite{ONS2023}, while the data for Paris is synthesized by Sebastian et al.~\cite{horl2021open} using the 2015 French population census provided by the National Institute of Statistics and Economic Studies~(INSEE)~\cite{INSEE2018}. Urban regions in Paris are defined at the commune level.
    
    \item \textbf{China.} We collect OD flow data for Beijing and Shanghai, two major metropolitan centers in China. The Beijing dataset is provided by a leading internet location service provider. The Shanghai data is extracted from Call Detail Records~(CDRs) by China's largest telecommunications company, following the method proposed by Iqbal et al.~\cite{iqbal2014development}. Urban regions in both cities are defined by subdistricts. All data was collected in 2016.
    
    \item \textbf{Brazil.} Julio et al.~\cite{chaves2023human} extracted OD flows for the Rio de Janeiro Metropolitan Area~(RJMA) in 2014 using CDRs. Urban regions are defined by Municípios.
    
    \item \textbf{Africa.} We use OD flow data extracted from CDRs collected in Senegal in 2013 as part of the D4D Challenge~\cite{de2014d4d}, processed following Iqbal et al.~\cite{iqbal2014development}. Urban regions are defined by Arrondissements. 
\end{itemize}

It is worth noting that real-world OD flow data is inherently difficult to collect. Datasets differ in sampling methods, temporal coverage, and spatial granularity, and often contain noise or bias. As such, they do not serve as perfect ground truth, but only as approximate references. In our evaluation, we focus on comparing the spatial patterns and distribution characteristics between generated and observed data, aiming to assess the plausibility and consistency of the generated flows under real-world constraints.

\begin{figure}[t]
    \centering
    \subfigure[Beijing~($\rho=$0.741)]{
        \includegraphics[width=0.3\textwidth]{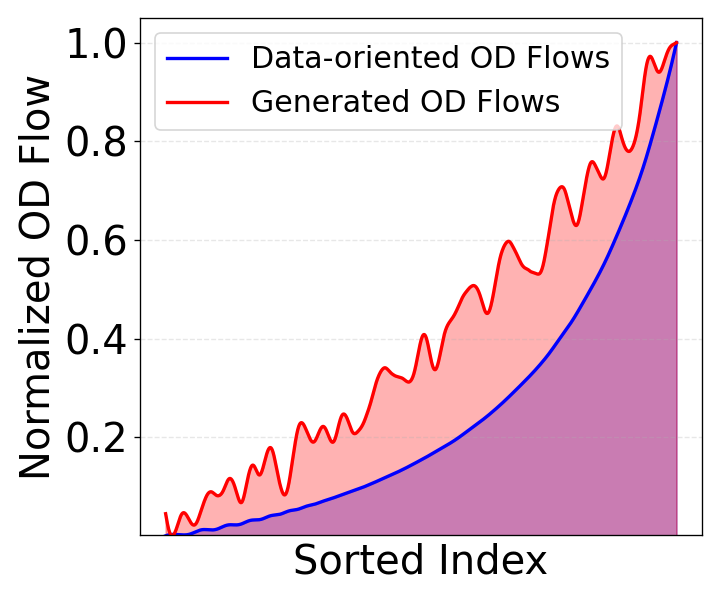}
    }
    \subfigure[Shanghai~($\rho=$0.623)]{
        \includegraphics[width=0.3\textwidth]{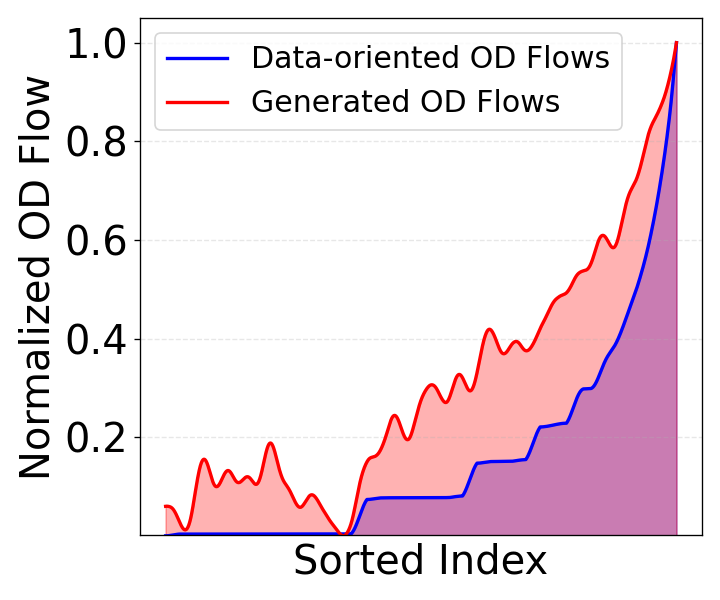}
    }
    \subfigure[London~($\rho=$0.361)]{
        \includegraphics[width=0.3\textwidth]{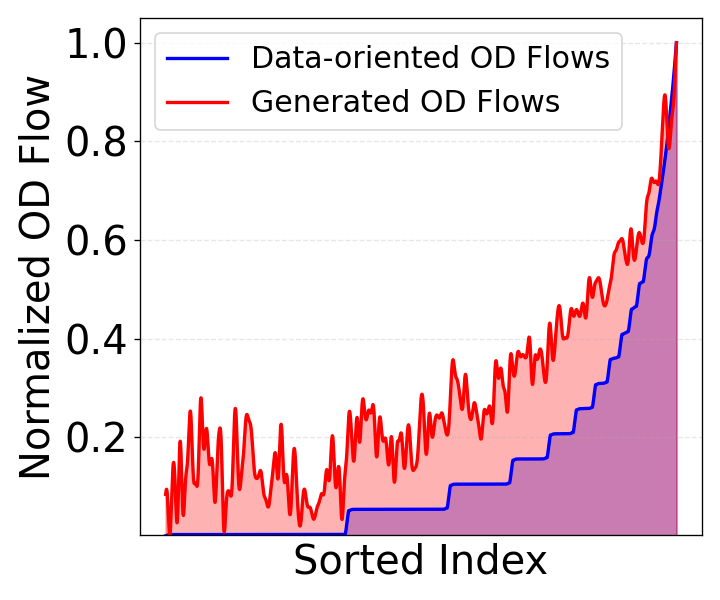}
    }
    \subfigure[Paris~($\rho=$0.465)]{
        \includegraphics[width=0.3\textwidth]{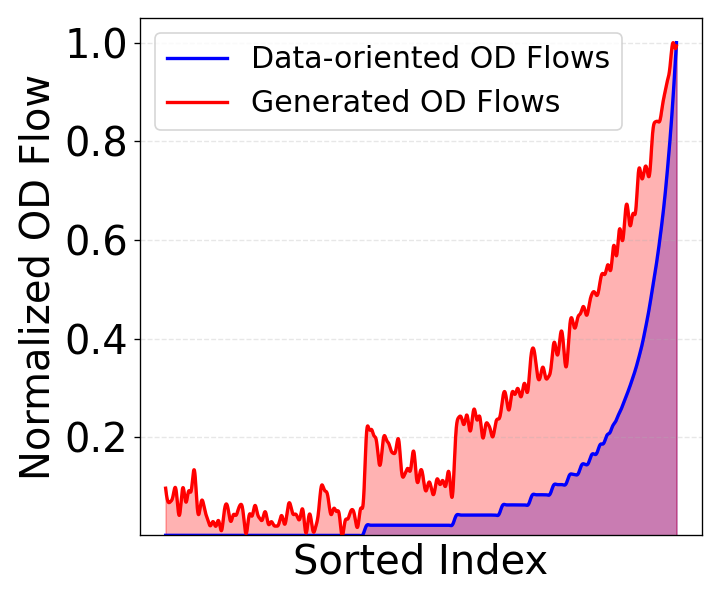}
    }
    \subfigure[Rio~($\rho=$0.816)]{
        \includegraphics[width=0.3\textwidth]{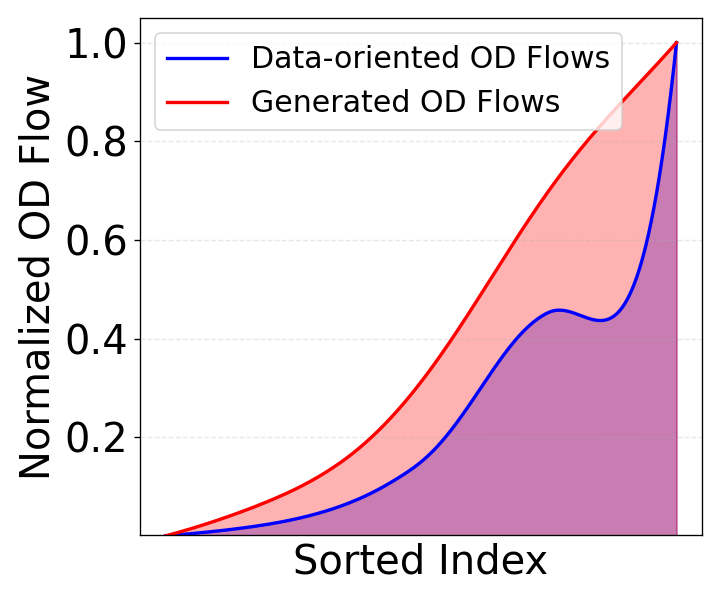}
    }
    \subfigure[Senegal~($\rho=$0.477)]{
        \includegraphics[width=0.3\textwidth]{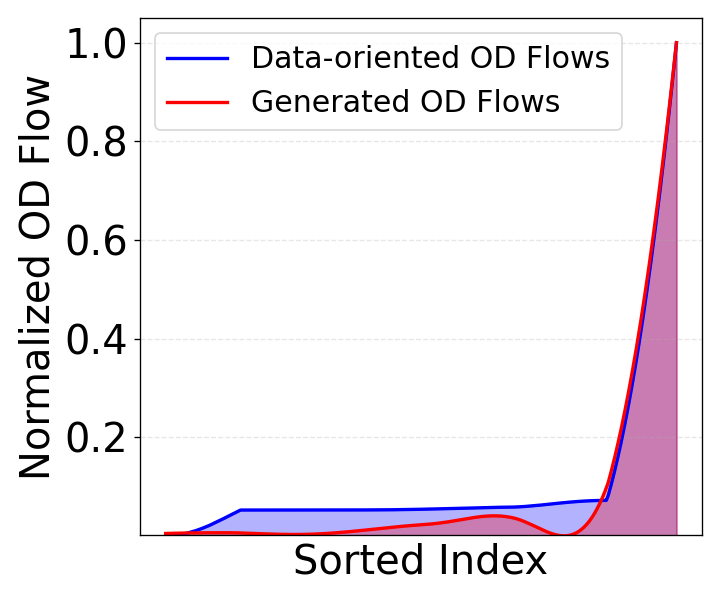}
    }
    \vspace{-0.3cm}
    \caption{Correlation analysis between generated OD flows and the flows extracted from diverse mobility-related data sources for typical urban areas around the world.}
    \label{fig:sort_line_cities}
    \vspace{-0.4cm}
\end{figure}
\vspace{-0.2cm}

\textbf{Experimental Results.} Based on the data collected from the representative urban areas above, we apply the proposed GlODGen framework to generate OD flows and compare the results with the corresponding real-world data. Given the heterogeneous data sources, varying sampling methods, and inherent noise across regions, a unified error-based evaluation metric is not applicable for direct comparison. To quantitatively assess the alignment between the generated and observed OD flows, we compute the Spearman rank correlation coefficients, visualized as rank-aligned normalized flow curve in Figure~\ref{fig:sort_line_cities}. Specifically, the data-oriented curve is constructed by the ascending ordered sort of the OD flow oriented from collected data, and the generated flow is also sorted in the same order. Then, the two curves are normalized to the same range and smoothed for better visualization. Although the distribution of the collected data is more skewed, the two curves are aligned with each other in a very similar manner, indicating that the generated OD flows are consistent with the real-world data in terms of data distribution. It is important to note that due to differences in data quality and sources, these correlation values are not directly comparable across cities. In addition, we qualitatively assess spatial patterns by visually comparing the generated and observed OD distributions on a case-by-case basis in Appendix~\ref{apdx:spatialvis}, under varying regional conditions, to provide a more intuitive understanding of the framework's performance. The experimental results show that the OD flows generated by GlODGen closely align with the real-world data in terms of distributional structure, as evidenced by the high Spearman rank correlations observed in most cases. Although lower correlations may arise due to limitations in the reference data itself, high correlation values provide strong empirical support for the credibility and applicability of the proposed framework.
\vspace{-0.4cm}
\section{Discussion} \label{sec:Discussion}
\vspace{-0.3cm}

\textbf{Conclusion} We find that the global publicly available data, i.e., satellite imagery and population data, has a strong representational power for profiling urban regions and supporting the generation of OD flows across diverse urban areas worldwide. Notably, public data can achieve 98\% of the performance of the traditional hard-to-obtain urban data. Building on this insight, we propose a novel data generator, GlODGen, which can generate realistic OD flows for any urban area worldwide. GlODGen integrates vision-language geo-foundation models to extract urban semantic features from satellite imagery and region-level population statistics and generates OD flows through a graph denoising diffusion-based approach. Extensive experiments across all over the world validate both the representational power of public data and the generalizability of GlODGen. For the convenience of practical use, we release GlODGen as an open-source tool, which can automatically complete the data acquisition, curation, preprocessing, urban semantic feature extraction, and OD flow generation with only the boundaries of urban regions given.

\textbf{Limitations and Boarder Impact} GlODGen currently has two main limitations. First, the OD flow generation model is primarily trained on U.S. data. While we demonstrate strong cross-continental generalization with the aid of satellite imagery, broader geographic coverage would improve robustness. Second, the generator produces only static commuting OD flows, limiting its applicability to dynamic, fine-grained tasks such as time-dependent traffic analysis. Despite these limitations, GlODGen can benefit various domains, including urban planning, transportation design, energy consumption, carbon emissions, and public health research.

\textbf{Ethical Claims} All individual-level location data used in this study have been rigorously anonymized. Additionally, all computations are conducted locally, with no reliance on cloud-based infrastructure, to guarantee both privacy and data security.

\clearpage


\bibliographystyle{plain}
\bibliography{reference}

\begin{thebibliography}{10}

\bibitem{abitbol2020interpretable}
Jacob~Levy Abitbol and Marton Karsai.
\newblock Interpretable socioeconomic status inference from aerial imagery through urban patterns.
\newblock {\em Nature Machine Intelligence}, 2(11):684--692, 2020.

\bibitem{ahn2023human}
Donghyun Ahn, Jeasurk Yang, Meeyoung Cha, Hyunjoo Yang, Jihee Kim, Sangyoon Park, Sungwon Han, Eunji Lee, Susang Lee, and Sungwon Park.
\newblock A human-machine collaborative approach measures economic development using satellite imagery.
\newblock {\em Nature Communications}, 14(1):6811, 2023.

\bibitem{axhausen2002observing}
Kay~W Axhausen, Andrea Zimmermann, Stefan Sch{\"o}nfelder, Guido Rindsf{\"u}ser, and Thomas Haupt.
\newblock Observing the rhythms of daily life: A six-week travel diary.
\newblock {\em Transportation}, 29(2):95--124, 2002.

\bibitem{ayush2020generating}
Kumar Ayush, Burak Uzkent, Marshall Burke, David Lobell, and Stefano Ermon.
\newblock Generating interpretable poverty maps using object detection in satellite images.
\newblock {\em arXiv preprint arXiv:2002.01612}, 2020.

\bibitem{ayush2021efficient}
Kumar Ayush, Burak Uzkent, Kumar Tanmay, Marshall Burke, David Lobell, and Stefano Ermon.
\newblock Efficient poverty mapping from high resolution remote sensing images.
\newblock In {\em Proceedings of the AAAI Conference on Artificial Intelligence}, volume~35, pages 12--20, 2021.

\bibitem{barbosa2018human}
Hugo Barbosa, Marc Barthelemy, Gourab Ghoshal, Charlotte~R James, Maxime Lenormand, Thomas Louail, Ronaldo Menezes, Jos{\'e}~J Ramasco, Filippo Simini, and Marcello Tomasini.
\newblock Human mobility: Models and applications.
\newblock {\em Physics Reports}, 734:1--74, 2018.

\bibitem{bergmeister2023efficient}
Andreas Bergmeister, Karolis Martinkus, Nathana{\"e}l Perraudin, and Roger Wattenhofer.
\newblock Efficient and scalable graph generation through iterative local expansion.
\newblock {\em arXiv preprint arXiv:2312.11529}, 2023.

\bibitem{bojchevski2018netgan}
Aleksandar Bojchevski, Oleksandr Shchur, Daniel Z{\"u}gner, and Stephan G{\"u}nnemann.
\newblock Netgan: Generating graphs via random walks.
\newblock In {\em International conference on machine learning}, pages 610--619. PMLR, 2018.

\bibitem{uscensusbureau2024}
U.S.~Census Bureau.
\newblock Lehd origin-destination employment statistics data (2002-2021), 2024.

\bibitem{burke2021using}
Marshall Burke, Anne Driscoll, David~B Lobell, and Stefano Ermon.
\newblock Using satellite imagery to understand and promote sustainable development.
\newblock {\em Science}, 371(6535):eabe8628, 2021.

\bibitem{calabrese2011estimating}
Francesco Calabrese, Giusy Di~Lorenzo, Liang Liu, and Carlo Ratti.
\newblock Estimating origin-destination flows using opportunistically collected mobile phone location data from one million users in boston metropolitan area.
\newblock 2011.

\bibitem{cats2014dynamic}
Oded Cats and Erik Jenelius.
\newblock Dynamic vulnerability analysis of public transport networks: mitigation effects of real-time information.
\newblock {\em Networks and Spatial Economics}, 14:435--463, 2014.

\bibitem{chaves2023human}
J{\'u}lio~C{\'e}sar Chaves, Moacyr~AHB da~Silva, Ricardo de~Souza~Alencar, Alexandre~G Evsukoff, and Vin{\'\i}cius da~Fonseca~Vieira.
\newblock Human mobility and socioeconomic datasets of the rio de janeiro metropolitan area.
\newblock {\em Data in brief}, 51:109695, 2023.

\bibitem{chen2017practical}
Jingbo Chen, Chengyi Wang, Dongxu He, Jiansheng Chen, and Anzhi Yue.
\newblock Practical bottom-up golf course detection using multispectral remote sensing imagery.
\newblock In {\em Image and Graphics: 9th International Conference, ICIG 2017, Shanghai, China, September 13-15, 2017, Revised Selected Papers, Part II 9}, pages 552--559. Springer, 2017.

\bibitem{chen2024characterizing}
Xiao-Jian Chen, Yuhui Zhao, Chaogui Kang, Xiaoyue Xing, Quanhua Dong, and Yu~Liu.
\newblock Characterizing the temporally stable structure of community evolution in intra-urban origin-destination networks.
\newblock {\em Cities}, 150:105033, 2024.

\bibitem{de2014d4d}
Yves-Alexandre de~Montjoye, Zbigniew Smoreda, Romain Trinquart, Cezary Ziemlicki, and Vincent~D Blondel.
\newblock D4d-senegal: the second mobile phone data for development challenge.
\newblock {\em arXiv preprint arXiv:1407.4885}, 2014.

\bibitem{ding2022don}
Yuxuan Ding, Lingqiao Liu, Chunna Tian, Jingyuan Yang, and Haoxuan Ding.
\newblock Don't stop learning: Towards continual learning for the clip model.
\newblock {\em arXiv preprint arXiv:2207.09248}, 2022.

\bibitem{dosovitskiy2020image}
Alexey Dosovitskiy, Lucas Beyer, Alexander Kolesnikov, Dirk Weissenborn, Xiaohua Zhai, Thomas Unterthiner, Mostafa Dehghani, Matthias Minderer, Georg Heigold, Sylvain Gelly, et~al.
\newblock An image is worth 16x16 words: Transformers for image recognition at scale.
\newblock {\em arXiv preprint arXiv:2010.11929}, 2020.

\bibitem{dwivedi2020generalization}
Vijay~Prakash Dwivedi and Xavier Bresson.
\newblock A generalization of transformer networks to graphs.
\newblock {\em arXiv preprint arXiv:2012.09699}, 2020.

\bibitem{feng2024citybench}
Jie Feng, Jun Zhang, Junbo Yan, Xin Zhang, Tianjian Ouyang, Tianhui Liu, Yuwei Du, Siqi Guo, and Yong Li.
\newblock Citybench: Evaluating the capabilities of large language model as world model.
\newblock {\em arXiv preprint arXiv:2406.13945}, 2024.

\bibitem{fu2024hyperbolic}
Xingcheng Fu, Yisen Gao, Yuecen Wei, Qingyun Sun, Hao Peng, Jianxin Li, and Xianxian Li.
\newblock Hyperbolic geometric latent diffusion model for graph generation.
\newblock In {\em Proceedings of the 41st International Conference on Machine Learning}, ICML'24. JMLR.org, 2024.

\bibitem{gundlegaard2016travel}
David Gundleg{\aa}rd, Clas Rydergren, Nils Breyer, and Botond Rajna.
\newblock Travel demand estimation and network assignment based on cellular network data.
\newblock {\em Computer Communications}, 95:29--42, 2016.

\bibitem{haefeli2022diffusion}
Kilian~Konstantin Haefeli, Karolis Martinkus, Nathana{\"e}l Perraudin, and Roger Wattenhofer.
\newblock Diffusion models for graphs benefit from discrete state spaces.
\newblock {\em arXiv preprint arXiv:2210.01549}, 2022.

\bibitem{han2020lightweight}
Sungwon Han, Donghyun Ahn, Hyunji Cha, Jeasurk Yang, Sungwon Park, and Meeyoung Cha.
\newblock Lightweight and robust representation of economic scales from satellite imagery.
\newblock In {\em Proceedings of the AAAI Conference on Artificial Intelligence}, volume~34, pages 428--436, 2020.

\bibitem{han2020learning}
Sungwon Han, Donghyun Ahn, Sungwon Park, Jeasurk Yang, Susang Lee, Jihee Kim, Hyunjoo Yang, Sangyoon Park, and Meeyoung Cha.
\newblock Learning to score economic development from satellite imagery.
\newblock In {\em Proceedings of the 26th ACM SIGKDD International Conference on Knowledge Discovery \& Data Mining}, pages 2970--2979, 2020.

\bibitem{hao2024urbanvlp}
Xixuan Hao, Wei Chen, Yibo Yan, Siru Zhong, Kun Wang, Qingsong Wen, and Yuxuan Liang.
\newblock Urbanvlp: A multi-granularity vision-language pre-trained foundation model for urban indicator prediction.
\newblock {\em arXiv preprint arXiv:2403.16831}, 2024.

\bibitem{hao2025urbanvlp}
Xixuan Hao, Wei Chen, Yibo Yan, Siru Zhong, Kun Wang, Qingsong Wen, and Yuxuan Liang.
\newblock Urbanvlp: Multi-granularity vision-language pretraining for urban socioeconomic indicator prediction.
\newblock In {\em Proceedings of the AAAI Conference on Artificial Intelligence}, volume~39, pages 28061--28069, 2025.

\bibitem{haraguchi2022human}
Masahiko Haraguchi, Akihiko Nishino, Akira Kodaka, Maura Allaire, Upmanu Lall, Liao Kuei-Hsien, Kaya Onda, Kota Tsubouchi, and Naohiko Kohtake.
\newblock Human mobility data and analysis for urban resilience: A systematic review.
\newblock {\em Environment and Planning B: Urban Analytics and City Science}, 49(5):1507--1535, 2022.

\bibitem{he2016deep}
Kaiming He, Xiangyu Zhang, Shaoqing Ren, and Jian Sun.
\newblock Deep residual learning for image recognition.
\newblock In {\em Proceedings of the IEEE conference on computer vision and pattern recognition}, pages 770--778, 2016.

\bibitem{he2018perceiving}
Zhiyuan He, Su~Yang, Weishan Zhang, and Jiulong Zhang.
\newblock Perceiving commerial activeness over satellite images.
\newblock In {\em Companion Proceedings of the The Web Conference 2018}, pages 387--394, 2018.

\bibitem{hong2020graph}
Danfeng Hong, Lianru Gao, Jing Yao, Bing Zhang, Antonio Plaza, and Jocelyn Chanussot.
\newblock Graph convolutional networks for hyperspectral image classification.
\newblock {\em IEEE Transactions on Geoscience and Remote Sensing}, 59(7):5966--5978, 2020.

\bibitem{hong20233d}
Yining Hong, Haoyu Zhen, Peihao Chen, Shuhong Zheng, Yilun Du, Zhenfang Chen, and Chuang Gan.
\newblock 3d-llm: Injecting the 3d world into large language models.
\newblock {\em Advances in Neural Information Processing Systems}, 36:20482--20494, 2023.

\bibitem{horl2021open}
Sebastian H{\"o}rl and Milos Balac.
\newblock Open synthetic travel demand for paris and {\^i}le-de-france: Inputs and output data.
\newblock {\em Data in Brief}, 39:107622, 2021.

\bibitem{iqbal2014development}
Md~Shahadat Iqbal, Charisma~F Choudhury, Pu~Wang, and Marta~C Gonz{\'a}lez.
\newblock Development of origin--destination matrices using mobile phone call data.
\newblock {\em Transportation Research Part C: Emerging Technologies}, 40:63--74, 2014.

\bibitem{jatavallabhula2023conceptfusion}
Krishna~Murthy Jatavallabhula, Alihusein Kuwajerwala, Qiao Gu, Mohd Omama, Tao Chen, Alaa Maalouf, Shuang Li, Ganesh Iyer, Soroush Saryazdi, Nikhil Keetha, et~al.
\newblock Conceptfusion: Open-set multimodal 3d mapping.
\newblock {\em arXiv preprint arXiv:2302.07241}, 2023.

\bibitem{jean2016combining}
Neal Jean, Marshall Burke, Michael Xie, W~Matthew Davis, David~B Lobell, and Stefano Ermon.
\newblock Combining satellite imagery and machine learning to predict poverty.
\newblock {\em Science}, 353(6301):790--794, 2016.

\bibitem{jia2020population}
Jayson~S Jia, Xin Lu, Yun Yuan, Ge~Xu, Jianmin Jia, and Nicholas~A Christakis.
\newblock Population flow drives spatio-temporal distribution of covid-19 in china.
\newblock {\em Nature}, 582(7812):389--394, 2020.

\bibitem{jo2022score}
Jaehyeong Jo, Seul Lee, and Sung~Ju Hwang.
\newblock Score-based generative modeling of graphs via the system of stochastic differential equations.
\newblock In {\em International conference on machine learning}, pages 10362--10383. PMLR, 2022.

\bibitem{li2020object}
Ke~Li, Gang Wan, Gong Cheng, Liqiu Meng, and Junwei Han.
\newblock Object detection in optical remote sensing images: A survey and a new benchmark.
\newblock {\em ISPRS journal of photogrammetry and remote sensing}, 159:296--307, 2020.

\bibitem{li2023learning}
Tong Li, Yanxin Xi, Huandong Wang, Yong Li, Sasu Tarkoma, and Pan Hui.
\newblock Learning representations of satellite imagery by leveraging point-of-interests.
\newblock {\em ACM Transactions on Intelligent Systems and Technology}, 14(4):1--32, 2023.

\bibitem{li2022predicting}
Tong Li, Shiduo Xin, Yanxin Xi, Sasu Tarkoma, Pan Hui, and Yong Li.
\newblock Predicting multi-level socioeconomic indicators from structural urban imagery.
\newblock In {\em Proceedings of the 31st ACM International Conference on Information \& Knowledge Management}, pages 3282--3291, 2022.

\bibitem{li2024vision}
Xiang Li, Congcong Wen, Yuan Hu, Zhenghang Yuan, and Xiao~Xiang Zhu.
\newblock Vision-language models in remote sensing: Current progress and future trends.
\newblock {\em IEEE Geoscience and Remote Sensing Magazine}, 2024.

\bibitem{liu2024remoteclip}
Fan Liu, Delong Chen, Zhangqingyun Guan, Xiaocong Zhou, Jiale Zhu, Qiaolin Ye, Liyong Fu, and Jun Zhou.
\newblock Remoteclip: A vision language foundation model for remote sensing.
\newblock {\em IEEE Transactions on Geoscience and Remote Sensing}, 2024.

\bibitem{liu2020learning}
Zhicheng Liu, Fabio Miranda, Weiting Xiong, Junyan Yang, Qiao Wang, and Claudio Silva.
\newblock Learning geo-contextual embeddings for commuting flow prediction.
\newblock In {\em Proceedings of the AAAI Conference on Artificial Intelligence}, volume~34, pages 808--816, 2020.

\bibitem{m2019semantic}
Rose M~Rustowicz, Robin Cheong, Lijing Wang, Stefano Ermon, Marshall Burke, and David Lobell.
\newblock Semantic segmentation of crop type in africa: A novel dataset and analysis of deep learning methods.
\newblock In {\em Proceedings of the IEEE/CVF Conference on Computer Vision and Pattern Recognition Workshops}, pages 75--82, 2019.

\bibitem{martinez2021fully}
Jorge Andres~Chamorro Martinez, Laura Elena~Cu{\'e} La~Rosa, Raul~Queiroz Feitosa, Ieda~Del’Arco Sanches, and Patrick~Nigri Happ.
\newblock Fully convolutional recurrent networks for multidate crop recognition from multitemporal image sequences.
\newblock {\em ISPRS Journal of Photogrammetry and Remote Sensing}, 171:188--201, 2021.

\bibitem{niu2020permutation}
Chenhao Niu, Yang Song, Jiaming Song, Shengjia Zhao, Aditya Grover, and Stefano Ermon.
\newblock Permutation invariant graph generation via score-based generative modeling.
\newblock In {\em International conference on artificial intelligence and statistics}, pages 4474--4484. PMLR, 2020.

\bibitem{ONS2023}
{Office for National Statistics}.
\newblock User guide to origin-destination data for census 2021, england and wales.
\newblock \url{https://www.ons.gov.uk}, October 2023.
\newblock Accessed: 2023-10-26.

\bibitem{pan2006cellular}
Changxuan Pan, Jiangang Lu, Shan Di, and Bin Ran.
\newblock Cellular-based data-extracting method for trip distribution.
\newblock {\em Transportation research record}, 1945(1):33--39, 2006.

\bibitem{perez2017poverty}
Anthony Perez, Christopher Yeh, George Azzari, Marshall Burke, David Lobell, and Stefano Ermon.
\newblock Poverty prediction with public landsat 7 satellite imagery and machine learning.
\newblock {\em arXiv preprint arXiv:1711.03654}, 2017.

\bibitem{pourebrahim2019trip}
Nastaran Pourebrahim, Selima Sultana, Amirreza Niakanlahiji, and Jean-Claude Thill.
\newblock Trip distribution modeling with twitter data.
\newblock {\em Computers, Environment and Urban Systems}, 77:101354, 2019.

\bibitem{ribeiro2019urban}
Paulo Jorge~Gomes Ribeiro and Lu{\'\i}s Ant{\'o}nio Pena~Jardim Gon{\c{c}}alves.
\newblock Urban resilience: A conceptual framework.
\newblock {\em Sustainable Cities and Society}, 50:101625, 2019.

\bibitem{robinson2018machine}
Caleb Robinson and Bistra Dilkina.
\newblock A machine learning approach to modeling human migration.
\newblock In {\em Proceedings of the 1st ACM SIGCAS Conference on Computing and Sustainable Societies}, pages 1--8, 2018.

\bibitem{rolf2021generalizable}
Esther Rolf, Jonathan Proctor, Tamma Carleton, Ian Bolliger, Vaishaal Shankar, Miyabi Ishihara, Benjamin Recht, and Solomon Hsiang.
\newblock A generalizable and accessible approach to machine learning with global satellite imagery.
\newblock {\em Nature communications}, 12(1):4392, 2021.

\bibitem{rong2023interdisciplinary}
Can Rong, Jingtao Ding, and Yong Li.
\newblock An interdisciplinary survey on origin-destination flows modeling: Theory and techniques.
\newblock {\em arXiv preprint arXiv:2306.10048}, 2023.

\bibitem{ronglarge}
Can Rong, Jingtao Ding, Yan Liu, and Yong Li.
\newblock A large-scale dataset and benchmark for commuting origin-destination flow generation.
\newblock In {\em The Thirteenth International Conference on Learning Representations}.

\bibitem{rong2023city}
Can Rong, Jingtao Ding, Zhicheng Liu, and Yong Li.
\newblock City-wide origin-destination matrix generation via graph denoising diffusion.
\newblock {\em arXiv preprint arXiv:2306.04873}, 2023.

\bibitem{rong2023goddag}
Can Rong, Jie Feng, and Jingtao Ding.
\newblock Goddag: Generating origin-destination flow for new cities via domain adversarial training.
\newblock {\em IEEE Transactions on Knowledge and Data Engineering}, 2023.

\bibitem{russwurm2020self}
Marc Ru{\ss}wurm and Marco K{\"o}rner.
\newblock Self-attention for raw optical satellite time series classification.
\newblock {\em ISPRS journal of photogrammetry and remote sensing}, 169:421--435, 2020.

\bibitem{shao2021one}
Erzhuo Shao, Jie Feng, Yingheng Wang, Tong Xia, and Yong Li.
\newblock One-shot transfer learning for population mapping.
\newblock In {\em Proceedings of the 30th ACM International Conference on Information \& Knowledge Management}, pages 1588--1597, 2021.

\bibitem{sharma2024kidsat}
Makkunda Sharma, Fan Yang, Duy-Nhat Vo, Esra Suel, Swapnil Mishra, Samir Bhatt, Oliver Fiala, William Rudgard, and Seth Flaxman.
\newblock Kidsat: satellite imagery to map childhood poverty dataset and benchmark.
\newblock {\em arXiv preprint arXiv:2407.05986}, 2024.

\bibitem{li2024m3luc}
Li~Sibo, Zhang Xin, Lin Yuming, and Li~Yong.
\newblock M3luc: Multi-modal model for urban land-use classification.
\newblock In {\em Proceedings of the 32nd ACM International Conference on Advances in Geographic Information Systems}, pages 1--12, 2024.

\bibitem{simini2021deep}
Filippo Simini, Gianni Barlacchi, Massimilano Luca, and Luca Pappalardo.
\newblock A deep gravity model for mobility flows generation.
\newblock {\em Nature communications}, 12(1):1--13, 2021.

\bibitem{simini2012universal}
Filippo Simini, Marta~C Gonz{\'a}lez, Amos Maritan, and Albert-L{\'a}szl{\'o} Barab{\'a}si.
\newblock A universal model for mobility and migration patterns.
\newblock {\em Nature}, 484(7392):96--100, 2012.

\bibitem{simonyan2014very}
Karen Simonyan and Andrew Zisserman.
\newblock Very deep convolutional networks for large-scale image recognition.
\newblock {\em arXiv preprint arXiv:1409.1556}, 2014.

\bibitem{sun2013understanding}
Lijun Sun, Kay~W Axhausen, Der-Horng Lee, and Xianfeng Huang.
\newblock Understanding metropolitan patterns of daily encounters.
\newblock {\em Proceedings of the National Academy of Sciences}, 110(34):13774--13779, 2013.

\bibitem{sun2021pbnet}
Xian Sun, Peijin Wang, Cheng Wang, Yingfei Liu, and Kun Fu.
\newblock Pbnet: Part-based convolutional neural network for complex composite object detection in remote sensing imagery.
\newblock {\em ISPRS Journal of Photogrammetry and Remote Sensing}, 173:50--65, 2021.

\bibitem{INSEE2018}
{The National Institute of Statistics and Economic Studies~(INSEE)}.
\newblock Individus localisés au canton-ou-ville en 2015.
\newblock \url{https://www.insee.fr/fr/statistiques/3625223}, 2015.

\bibitem{uzkent2019learning}
Burak Uzkent, Evan Sheehan, Chenlin Meng, Zhongyi Tang, Marshall Burke, David Lobell, and Stefano Ermon.
\newblock Learning to interpret satellite images using wikipedia.
\newblock In {\em Proceedings of the Twenty-Eighth International Joint Conference on Artificial Intelligence}, 2019.

\bibitem{vignac2022digress}
Clement Vignac, Igor Krawczuk, Antoine Siraudin, Bohan Wang, Volkan Cevher, and Pascal Frossard.
\newblock Digress: Discrete denoising diffusion for graph generation.
\newblock {\em arXiv preprint arXiv:2209.14734}, 2022.

\bibitem{wang2018deep}
Anna~X Wang, Caelin Tran, Nikhil Desai, David Lobell, and Stefano Ermon.
\newblock Deep transfer learning for crop yield prediction with remote sensing data.
\newblock In {\em Proceedings of the 1st ACM SIGCAS Conference on Computing and Sustainable Societies}, pages 1--5, 2018.

\bibitem{xi2022beyond}
Yanxin Xi, Tong Li, Huandong Wang, Yong Li, Sasu Tarkoma, and Pan Hui.
\newblock Beyond the first law of geography: Learning representations of satellite imagery by leveraging point-of-interests.
\newblock In {\em Proceedings of the ACM Web Conference 2022}, pages 3308--3316, 2022.

\bibitem{xi2023satellite}
Yanxin Xi, Yu~Liu, Tong Li, Jintao Ding, Yunke Zhang, Sasu Tarkoma, Yong Li, and Pan Hui.
\newblock A satellite imagery dataset for long-term sustainable development in united states cities.
\newblock {\em Scientific data}, 10(1):866, 2023.

\bibitem{xi2024pixels}
Yanxin Xi, Yu~Liu, Zhicheng Liu, Sasu Tarkoma, Pan Hui, and Yong Li.
\newblock From pixels to progress: generating road network from satellite imagery for socioeconomic insights in impoverished areas.
\newblock In {\em Proceedings of the Thirty-Third International Joint Conference on Artificial Intelligence}, IJCAI '24, 2024.

\bibitem{xia2018dota}
Gui-Song Xia, Xiang Bai, Jian Ding, Zhen Zhu, Serge Belongie, Jiebo Luo, Mihai Datcu, Marcello Pelillo, and Liangpei Zhang.
\newblock Dota: A large-scale dataset for object detection in aerial images.
\newblock In {\em Proceedings of the IEEE conference on computer vision and pattern recognition}, pages 3974--3983, 2018.

\bibitem{xiao2024refound}
Congxi Xiao, Jingbo Zhou, Yixiong Xiao, Jizhou Huang, and Hui Xiong.
\newblock Refound: Crafting a foundation model for urban region understanding upon language and visual foundations.
\newblock In {\em Proceedings of the 30th ACM SIGKDD Conference on Knowledge Discovery and Data Mining}, pages 3527--3538, 2024.

\bibitem{yan2024urbanclip}
Yibo Yan, Haomin Wen, Siru Zhong, Wei Chen, Haodong Chen, Qingsong Wen, Roger Zimmermann, and Yuxuan Liang.
\newblock Urbanclip: Learning text-enhanced urban region profiling with contrastive language-image pretraining from the web.
\newblock In {\em Proceedings of the ACM on Web Conference 2024}, pages 4006--4017, 2024.

\bibitem{yeh2021sustainbench}
Christopher Yeh, Chenlin Meng, Sherrie Wang, Anne Driscoll, Erik Rozi, Patrick Liu, Jihyeon Lee, Marshall Burke, David~B Lobell, and Stefano Ermon.
\newblock Sustainbench: Benchmarks for monitoring the sustainable development goals with machine learning.
\newblock {\em arXiv preprint arXiv:2111.04724}, 2021.

\bibitem{yeh2020using}
Christopher Yeh, Anthony Perez, Anne Driscoll, George Azzari, Zhongyi Tang, David Lobell, Stefano Ermon, and Marshall Burke.
\newblock Using publicly available satellite imagery and deep learning to understand economic well-being in africa.
\newblock {\em Nature communications}, 11(1):2583, 2020.

\bibitem{you2017deep}
Jiaxuan You, Xiaocheng Li, Melvin Low, David Lobell, and Stefano Ermon.
\newblock Deep gaussian process for crop yield prediction based on remote sensing data.
\newblock In {\em Proceedings of the AAAI conference on artificial intelligence}, volume~31, 2017.

\bibitem{zeng2024carbon}
Jinwei Zeng, Yu~Liu, Jingtao Ding, Jian Yuan, and Yong Li.
\newblock Estimating on-road transportation carbon emissions from open data of road network and origin-destination flow data.
\newblock In {\em Proceedings of the AAAI Conference on Artificial Intelligence}, volume~37, pages 14539--14547, 2024.

\bibitem{zhang2024uv}
Xin Zhang, Yu~Liu, Yuming Lin, Qingmin Liao, and Yong Li.
\newblock Uv-sam: Adapting segment anything model for urban village identification.
\newblock {\em arXiv preprint arXiv:2401.08083}, 2024.

\bibitem{zhong2024urbancross}
Siru Zhong, Xixuan Hao, Yibo Yan, Ying Zhang, Yangqiu Song, and Yuxuan Liang.
\newblock Urbancross: enhancing satellite image-text retrieval with cross-domain adaptation.
\newblock In {\em Proceedings of the 32nd ACM International Conference on Multimedia}, pages 6307--6315, 2024.

\bibitem{zhou2024towards}
Yue Zhou, Litong Feng, Yiping Ke, Xue Jiang, Junchi Yan, Xue Yang, and Wayne Zhang.
\newblock Towards vision-language geo-foundation model: A survey.
\newblock {\em arXiv preprint arXiv:2406.09385}, 2024.

\bibitem{zipf1946p}
George~Kingsley Zipf.
\newblock The p 1 p 2/d hypothesis: on the intercity movement of persons.
\newblock {\em American sociological review}, 11(6):677--686, 1946.

\end{thebibliography}

\clearpage

\section*{NeurIPS Paper Checklist}

\begin{enumerate}

    \item {\bf Claims}
        \item[] Question: Do the main claims made in the abstract and introduction accurately reflect the paper's contributions and scope?
        \item[] Answer: \answerYes{} 
        \item[] Justification: See Abstract and Section~\ref{sec:Introduction}.
    
    \item {\bf Limitations}
        \item[] Question: Does the paper discuss the limitations of the work performed by the authors?
        \item[] Answer: \answerYes{} 
        \item[] Justification: See Section~\ref{sec:Discussion}.
    
    \item {\bf Theory assumptions and proofs}
        \item[] Question: For each theoretical result, does the paper provide the full set of assumptions and a complete (and correct) proof?
        \item[] Answer: \answerNA{} 
        \item[] Justification: Our work is not about theoretical results.
    
        \item {\bf Experimental result reproducibility}
        \item[] Question: Does the paper fully disclose all the information needed to reproduce the main experimental results of the paper to the extent that it affects the main claims and/or conclusions of the paper (regardless of whether the code and data are provided or not)?
        \item[] Answer: \answerYes{} 
        \item[] Justification: See Section~\ref{sec:experiments} and Github link provided in the paper.

    \item {\bf Open access to data and code}
        \item[] Question: Does the paper provide open access to the data and code, with sufficient instructions to faithfully reproduce the main experimental results, as described in supplemental material?
        \item[] Answer: \answerYes{} 
        \item[] Justification: See Github link provided in the paper.

    \item {\bf Experimental setting/details}
        \item[] Question: Does the paper specify all the training and test details (e.g., data splits, hyperparameters, how they were chosen, type of optimizer, etc.) necessary to understand the results?
        \item[] Answer: \answerYes{} 
        \item[] Justification: See Section~\ref{sec:experiments}.
    
    \item {\bf Experiment statistical significance}
        \item[] Question: Does the paper report error bars suitably and correctly defined or other appropriate information about the statistical significance of the experiments?
        \item[] Answer: \answerYes{} 
        \item[] Justification: See Section~\ref{sec:experiments}.
    
    \item {\bf Experiments compute resources}
        \item[] Question: For each experiment, does the paper provide sufficient information on the computer resources (type of compute workers, memory, time of execution) needed to reproduce the experiments?
        \item[] Answer: \answerYes{} 
        \item[] Justification: See Section~\ref{sec:experiments}.
        
    \item {\bf Code of ethics}
        \item[] Question: Does the research conducted in the paper conform, in every respect, with the NeurIPS Code of Ethics \url{https://neurips.cc/public/EthicsGuidelines}?
        \item[] Answer: \answerYes{} 
        \item[] Justification: See Section~\ref{sec:Discussion}.

    \item {\bf Broader impacts}
        \item[] Question: Does the paper discuss both potential positive societal impacts and negative societal impacts of the work performed?
        \item[] Answer: \answerYes{} 
        \item[] Justification: See Section~\ref{sec:Discussion}.
        
    \item {\bf Safeguards}
        \item[] Question: Does the paper describe safeguards that have been put in place for responsible release of data or models that have a high risk for misuse (e.g., pretrained language models, image generators, or scraped datasets)?
        \item[] Answer: \answerYes{} 
        \item[] Justification: See Section~\ref{sec:Discussion}.
    
    \item {\bf Licenses for existing assets}
        \item[] Question: Are the creators or original owners of assets (e.g., code, data, models), used in the paper, properly credited and are the license and terms of use explicitly mentioned and properly respected?
        \item[] Answer: \answerYes{} 
        \item[] Justification: We have provided the license information in the Github repository.
    
    \item {\bf New assets}
        \item[] Question: Are new assets introduced in the paper well documented and is the documentation provided alongside the assets?
        \item[] Answer: \answerYes{} 
        \item[] Justification: See Section~\ref{sec:experiments}.
    
    \item {\bf Crowdsourcing and research with human subjects}
        \item[] Question: For crowdsourcing experiments and research with human subjects, does the paper include the full text of instructions given to participants and screenshots, if applicable, as well as details about compensation (if any)? 
        \item[] Answer: \answerNA{} 
        \item[] Justification: Our work does not involve crowdsourcing nor research with human subjects.
    
    \item {\bf Institutional review board (IRB) approvals or equivalent for research with human subjects}
        \item[] Question: Does the paper describe potential risks incurred by study participants, whether such risks were disclosed to the subjects, and whether Institutional Review Board (IRB) approvals (or an equivalent approval/review based on the requirements of your country or institution) were obtained?
        \item[] Answer: \answerNA{} 
        \item[] Justification: Our work does not involve crowdsourcing nor research with human subjects.
    
    \item {\bf Declaration of LLM usage}
        \item[] Question: Does the paper describe the usage of LLMs if it is an important, original, or non-standard component of the core methods in this research? Note that if the LLM is used only for writing, editing, or formatting purposes and does not impact the core methodology, scientific rigorousness, or originality of the research, declaration is not required.
        \item[] Answer: \answerYes{} 
        \item[] Justification: We use LLMs for checking the writing errors in the paper.
    
    \end{enumerate}

\appendix

\section{Additional Details of GlODGen} \label{apdx:glodgen}
\subsection{Forward Diffusion Process}

We give an introduction to the OD generation model based on graph diffusion in this part. Like the traditional diffusion model, graph diffusion-based models include two main processes: the forward diffusion and the reverse denoising process. During training, the forward diffusion process is utilized to create the training dataset for the denoising network. This process involves gradually adding small Gaussian noise to the original OD flow data, eventually transforming them into pure noise that adheres to a standard Gaussian distribution, as illustrated in the formula below:
\begin{equation}
    \begin{split}
        \begin{aligned}
        & q(F_{ij}^t|F_{ij}^{t-1})=\mathcal{N}(F_{ij}^t; \sqrt{1-\beta_t} F_{ij}^{t-1} , \beta_t \mathbf{I}),\\
        & q(F_{ij}^1, ..., F_{ij}^T|F_{ij}^0) = \prod_{t=1}^T {q(F_{ij}^t|q^{t-1})},
        \end{aligned}
    \end{split}
\end{equation}
where $F_{ij}^t$ denotes the flow starting at $r_i$ flowing into $r_j$ at $t$-th diffusion step, $\mathcal{N}$ is the Gaussian distribution, $\beta_t$ is the noise level at time $t$, and $\mathbf{I}$ is the identity matrix.

The denoising process works in reverse of the forward diffusion process, which is already described in the main paper.

\subsection{Training}

The OD flow generation model requires training on diverse datasets to generalize effectively across global scales. We employ data from the whole United States to expose the model to varied mobility patterns, including those from developed and underdeveloped areas.

To generate training samples, the forward process is employed to produce OD flows with varying degrees of noise. Node inputs consist of semantic features extracted from satellite images and population data, while edge inputs are formed by noisy OD flows sampled from the forward process. The model is designed to predict the Gaussian noise disturbances to be removed at each noise level. The training employs the mean squared error~(MSE) as the loss function, and optimization is conducted using the Adam optimizer, consistent with~\cite{rong2023city}. The loss function is formulated as follows:
\begin{equation}
    \begin{split}
    \begin{aligned}
        \mathcal{L} = \mathbb{E}_{t,\epsilon \sim \mathcal{N}(0,\mathbf{I})} \left[ \| \epsilon - \epsilon_\theta(\mathbf{F}^t,t,\mathcal{C_\mathcal{R}}) \|_2^2 \right]
    \end{aligned}
    \end{split}
\end{equation}
where $\| \dot \|$ denotes the $L-2$ norm.

\section{Experimental Details}

\subsection{Detailed Introduction of Existing OD Flow Generation Models} \label{apdx:modelintro}

\begin{itemize}
    \item \textbf{Random Forest.~\cite{pourebrahim2019trip}} The Random Forest model is a type of ensemble learning method that builds multiple decision trees and combines their outputs to make a final prediction. It is a popular choice for its simplicity and effectiveness in handling complex relationships between various urban indicators profiling urban regions and OD flows.
    \item \textbf{DeepGravity.~\cite{simini2021deep}} DeepGravity is a multi-layer perceptron~(MLP)-based model inspired by the traditional gravity model. It models the process of decision making of destination choice as a classification problem and uses the softmax function to calculate the probability of each destination. With the outflow given, the model can generate the OD flows by multiplying the probability of each destination and the outflow. In this paper, we directly generate the OD flows by using the model because there is no outflow for any urban areas around the world.
    \item \textbf{GMEL.~\cite{liu2020learning}} GMEL is a graph-based model that uses the graph structure of the urban regions to generate the OD flows. It models the urban regions by aggregating the features of the regions and the features of the connections between the regions. After the geo-contextual embedding learned from the model, random forest is used to generate the OD flows based on the urban region embeddings.
    \item \textbf{NetGAN.~\cite{bojchevski2018netgan}} NetGAN is a generative adversarial network~(GAN)-based model that minimizes the Wasserstein distance between the random walk sequences of the generated and real OD flows, where the mobility flow networks are represented as graphs. The original NetGAN model is designed for unweighted networks. We adapt it to weighted networks by using the weighted adjacency matrix of the urban regions. The weighted matrix is the OD matrix of the city.
    \item \textbf{WEDAN.~\cite{ronglarge}} WEDAN is a graph denoising diffusion-based model that uses the graph structure of the urban regions to generate the OD flows. It models the urban regions as nodes and OD flows as the directed weighted edges. The model uses the conditional graph denoising diffusion process to generate the OD flows given the urban region features of the city.
\end{itemize}

\subsection{Details of Evaluation Metrics} \label{apdx:evalmetric}

For evaluation, we adopt root mean square error (RMSE), normalized RMSE (NRMSE), and common part of commuting (CPC) as metrics, with the computations detailed below:
\begin{equation}
    \begin{split}
    \begin{aligned}
        & \mathrm{RMSE} = \sqrt{\frac{1}{|\textbf{F}|} \sum\nolimits_{r_i,r_j \in \mathcal{R}} ||\textbf{F}_{ij} - \hat{\textbf{F}}_{ij}||_2^2}, \\
        & \mathrm{NRMSE} = \frac{\mathrm{RMSE}}{\sqrt{\frac{1}{N^2} \sum\nolimits_{r_i, r_j \in \mathcal{R}} || F_{ij} - \bar{F}_{ij} ||_2^2}}, \\
        & \mathrm{CPC} = \frac{2 \sum_{r_i, r_j \in \mathcal{R}} \min(\mathbf{F}_{ij}, \hat{\mathbf{F}}_{ij})}{\sum_{r_i, r_j \in \mathcal{R}} \mathbf{F}_{ij} + \sum_{r_i, r_j \in \mathcal{R}} \hat{\mathbf{F}}_{ij}}
    \end{aligned}
    \end{split}
\end{equation}
where $\bar{\mathbf{F}}$ represents the expectation of the OD flows $\mathbf{F}$, which is extracted from collected data.

\subsection{Data Processing for Typical Urban Areas Around the World} \label{apdx:datadoc}

\begin{itemize}
    \item \textbf{United States} OD flows of urban areas in the United States are collected and provided by the National Census Bureau through the Longitudinal Employer-Household Dynamics Origin-Destination Employment Statistics~(LODES)~\cite{uscensusbureau2024}. This dataset provides commuting flows across all census blocks. In line with previous studies~\cite{pourebrahim2019trip,liu2020learning,robinson2018machine,simini2021deep}, we aggregate this data to the census tract level, which defines the regions in our work, while counties represent areas. This dataset offers extensive population coverage and high accuracy, making it widely used in research related to human mobility. In this work, we utilized data collected in 2018.
    \item \textbf{China.} OD flow data are collected for Beijing and Shanghai, two key cities in China. The OD flow data for Beijing is provided by a major internet location service provider in China. Data from Shanghai are extracted from CNAs, provided by China's largest telecommunications company, using the method proposed by Iqbal et al.~\cite{iqbal2014development}.
    \item \textbf{Brazil.} Julio et al.~\cite{chaves2023human} utilized CDRs to extract OD flows of the Rio de Janeiro Metropolitan Area~(RJMA) in 2014. Urban regions in RJMA are defined by Municípios. This dataset provides total flows including repeated records of the same individuals. Therefore, there is a certain degree of bias.
    \item \textbf{Africa.} We identified the dataset of CDRs within Senegal in 2013~\cite{de2014d4d}. The data was processed to extract OD flows using the method proposed by Iqbal et al.~\cite{iqbal2014development}. Urban regions in Senegal are defined by Arrondissements. In Africa, the collection of mobility-related data is much more challenging due to the lack of comprehensive and up-to-date infrastructure and limited technological resources, and funding of large-scale data gathering efforts. Because of the sparsity and scarcity, the spatial granularity of the data cannot be further refined.
\end{itemize}

\subsection{Spatial Visualization of Compraison between Generated and Collected OD Flows} \label{apdx:spatialvis}

We would like to clarify that some visualizations for Beijing, Shanghai, Rio de Janeiro, and a region in Senegal also \textbf{appear in a separate submission currently under review at Scientific Data}. In that paper, we present a large-scale dataset of commuting OD flows for 1,625 global cities, generated using the data generator GlODGen introduced in this work, to support research on sustainable urban development.

The core contributions and goals of the two submissions are fundamentally different:
\begin{itemize}[leftmargin=*]
    \item The Scientific Data submission focuses on the global-scale dataset as a scientific contribution.
    \item The present paper focuses on the automated generator~(GlODGen) and its methodology for worldwide OD flow generation, which highlights the potential of generalization for urban areas around the world.
\end{itemize}
Although these cities appear in both papers, the visualizations are styled differently and are used for distinct purposes: In Scientific Data, they are presented to validate the quality and scope of the dataset, whereas, in this work, they demonstrate the data generator's generalizability and applicability across diverse global cities. We emphasize that the experimental results are not reused, but rather, the same generated data is used in different contexts to support complementary contributions in the two submissions.

The experimental results of spatial visualization are shown in the following.

To demonstrate the potential of our framework, which relies solely on public data, for generating OD flows globally, we conduct experiments in urban areas including the representative cities in the United States, Europe, China, Brazil, and Africa. Since OD flow data of these diverse areas are collected from different data sources with different sampling biases and noise, we cannot use a unified standard to evaluate the generation of these areas and compare generated OD flows with flows from data in terms of traditional error-based metrics. We visually and qualitatively compare the spatial distribution of the generation and data in a case-by-case manner under different scenarios and data conditions to provide an intuitive understanding of the performance of our framework.

\subsubsection{China} \label{sec:exp_china}

\begin{figure}[h]
    \centering
    \subfigure[Generated OD flows in Beijing]{
        \includegraphics[width=0.45\linewidth]{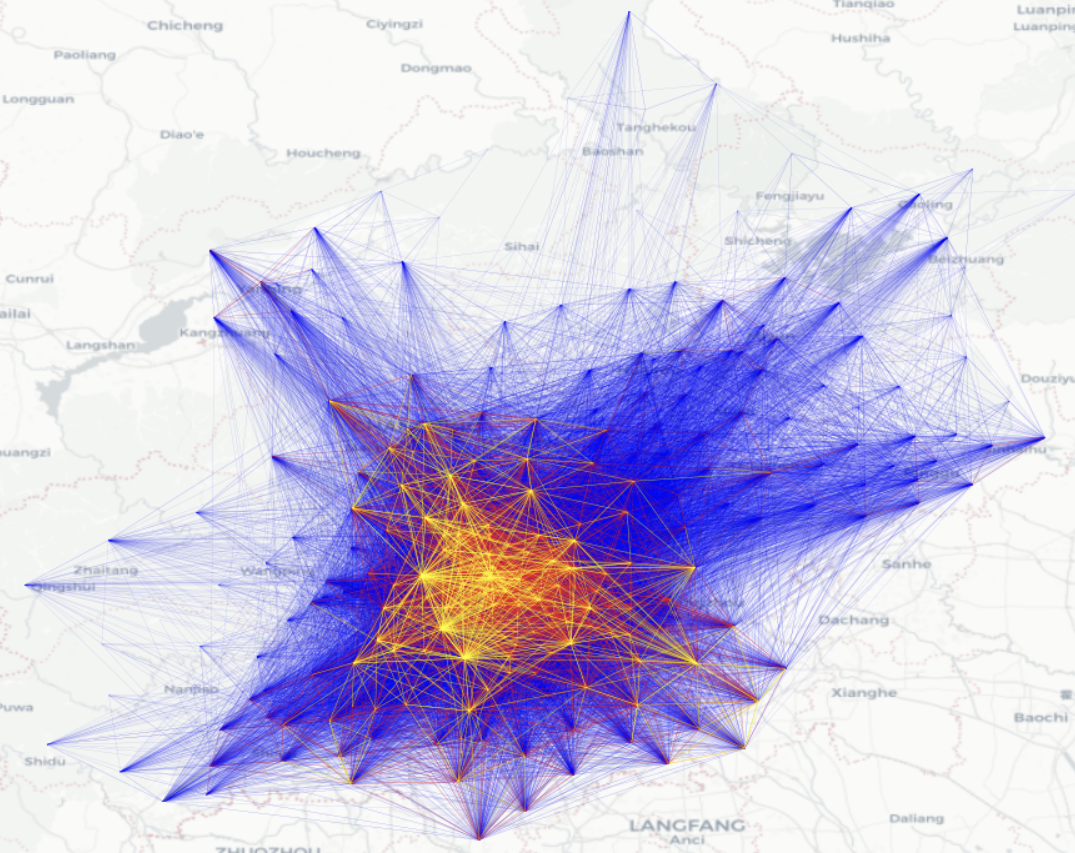}
    }
    \subfigure[Data-oriented OD flows in Beijing]{
        \includegraphics[width=0.45\linewidth]{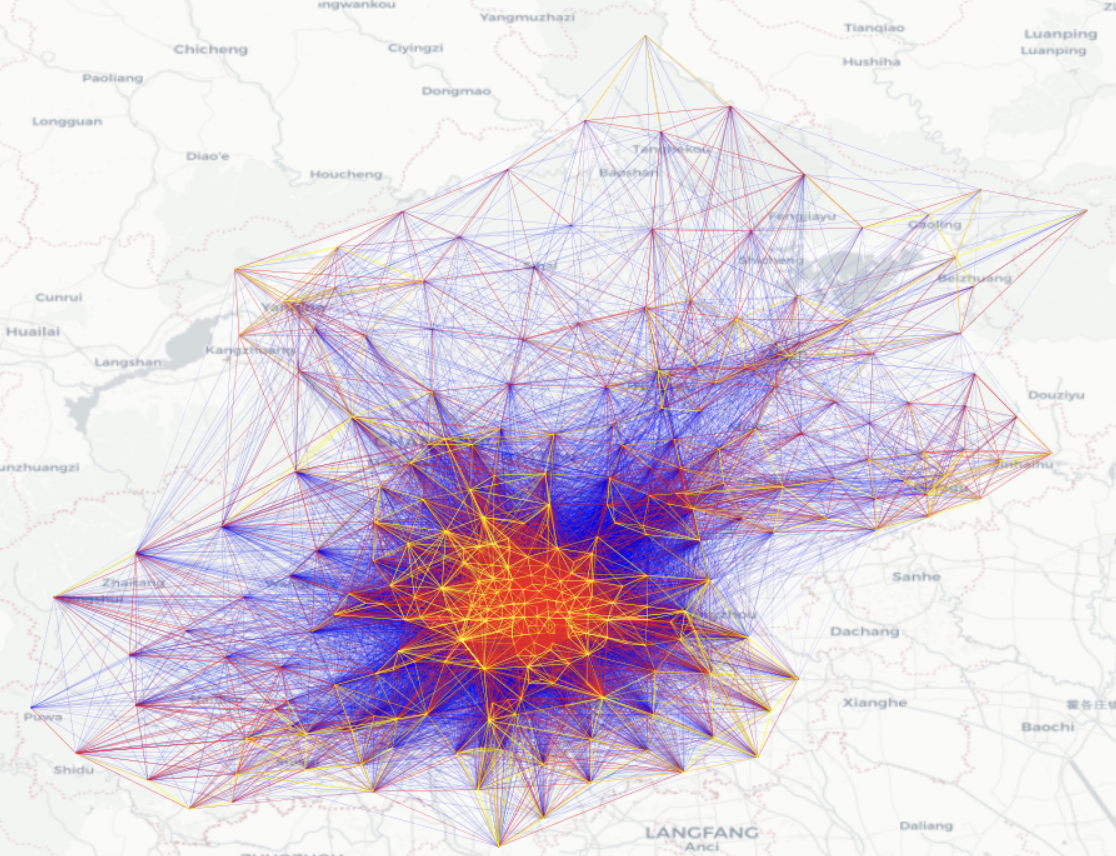}
    }
    \subfigure[Generated OD flows in Shanghai]{
        \includegraphics[width=0.45\linewidth]{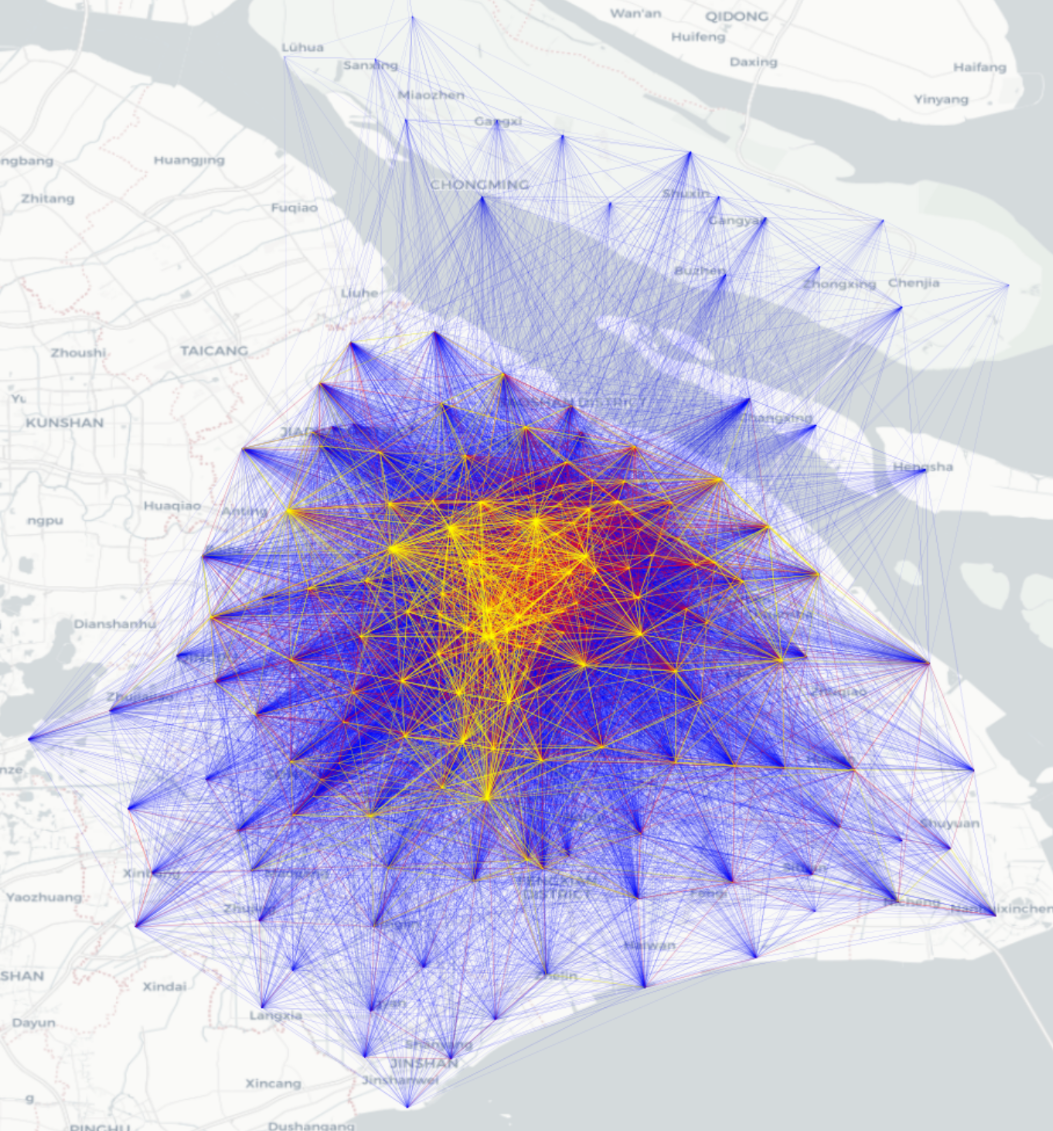}
    }
    \subfigure[Data-oriented OD flows in Shanghai]{
        \includegraphics[width=0.45\linewidth]{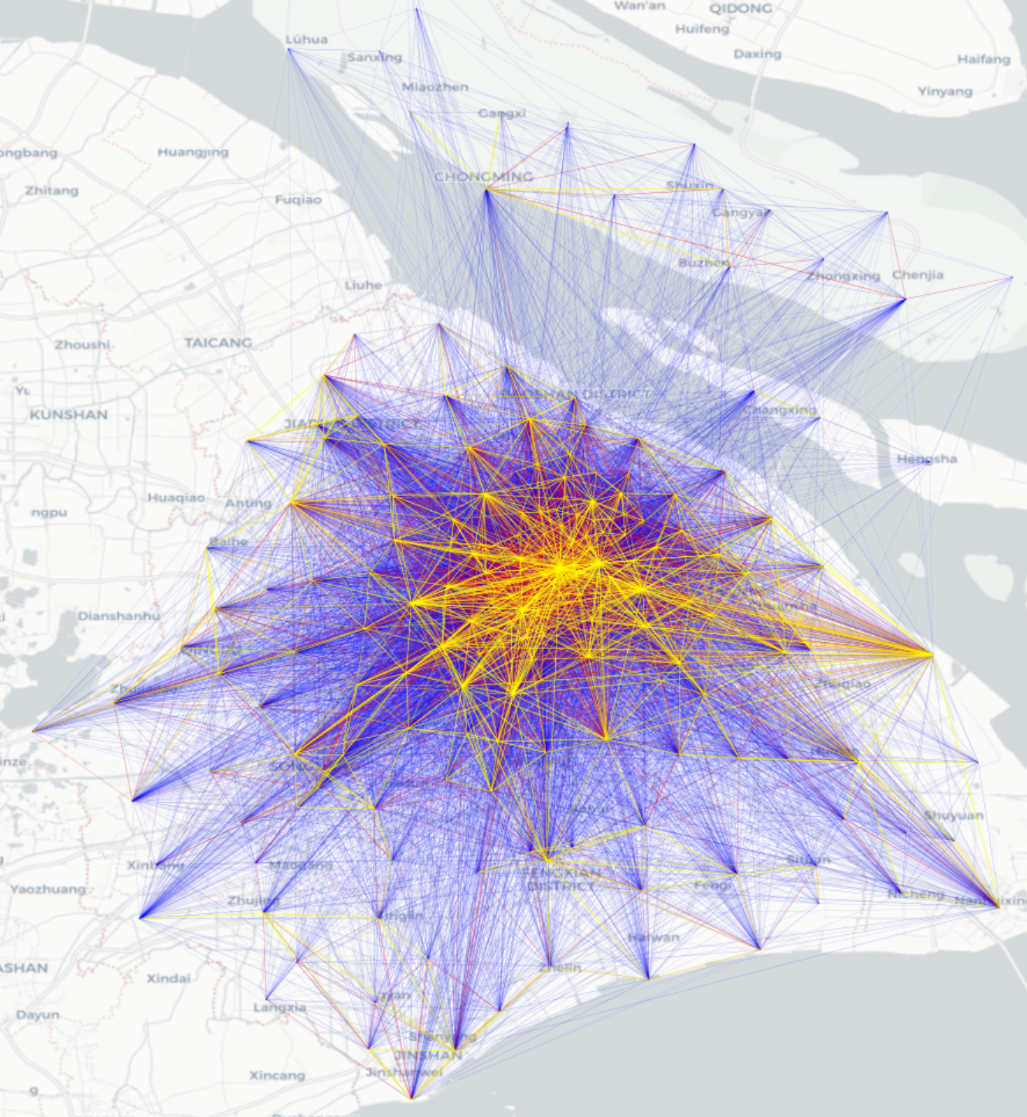}
    }
    \caption{Visualization of the generated and data-oriented OD flows in Beijing and Shanghai. The generated OD flows show high consistency with the data-oriented data in both cities, demonstrating the effectiveness of GlODGen in capturing real-world mobility patterns.}
    \label{fig:spviz_China}
\end{figure}

We generate OD flows for Beijing and Shanghai, two representative cities in China. The visually qualitative comparisons of the generation and data are shown in Figure~\ref{fig:spviz_China}. Figures from above to below are Beijing and Shanghai respectively and the left column is the generation, while the right column is OD flows from data. The generated OD flows exhibit a notable spatial similarity to the data, with the city center and boundary shapes accurately reflected in their spatial distribution. The agreement between generated flows and observed data under the given sampling scheme highlights the framework's validity. Nonetheless, slight discrepancies exist. For instance, flows from Beijing's suburbs to the city center are underrepresented, and generated flows for Shanghai's outskirts are somewhat denser than observed. These differences could stem from the unique urban characteristics of large cities, in which global public data may not fully follow a uniform regularity.

\subsubsection{Europe} \label{sec:exp_europe}

\begin{figure}[h]
    \centering
    \subfigure[Generated OD flows in London]{
        \includegraphics[width=0.45\linewidth]{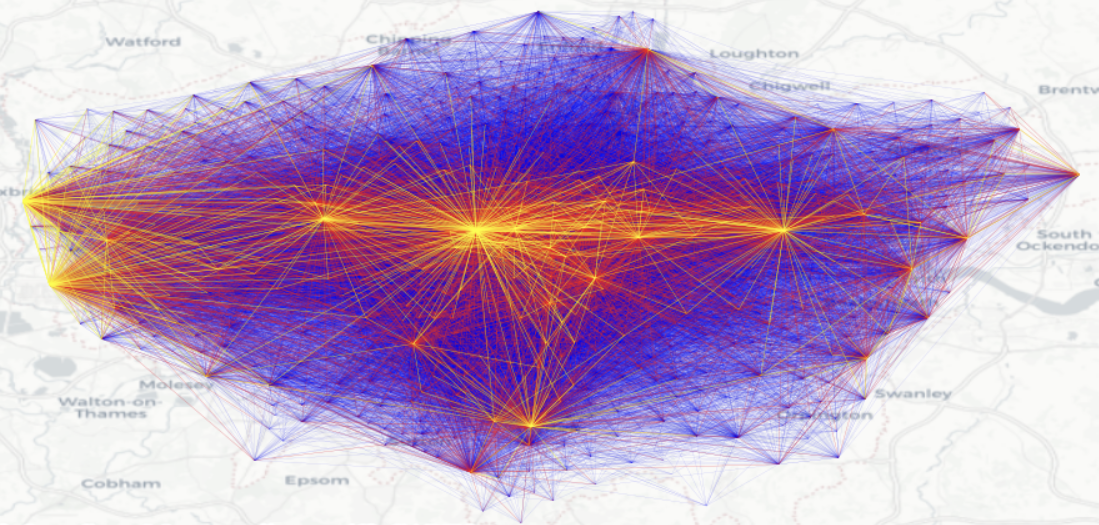}
    }
    \subfigure[Data-oriented OD flows in London]{
        \includegraphics[width=0.45\linewidth]{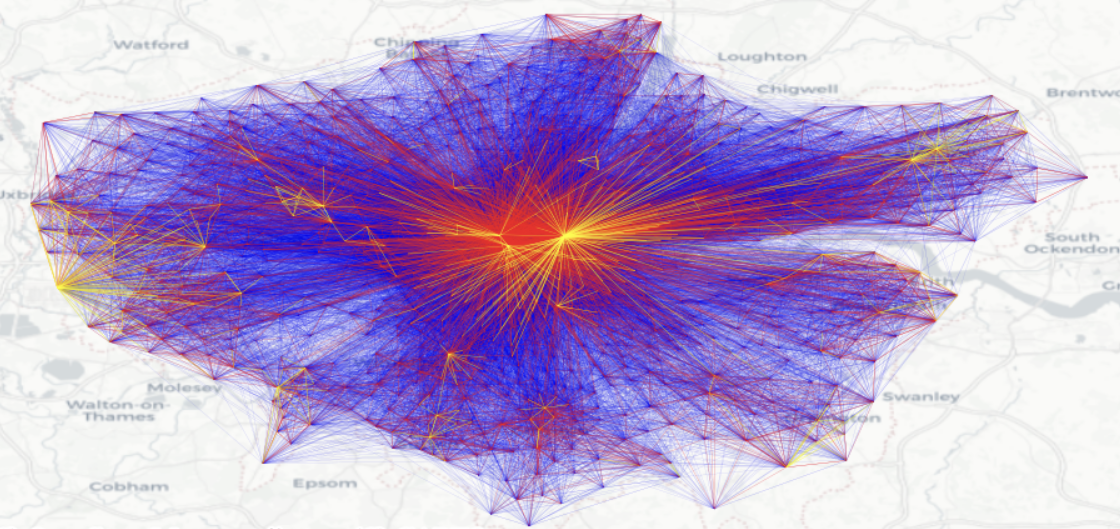}
    }
    \subfigure[Generated OD flows in Paris]{
        \includegraphics[width=0.45\linewidth]{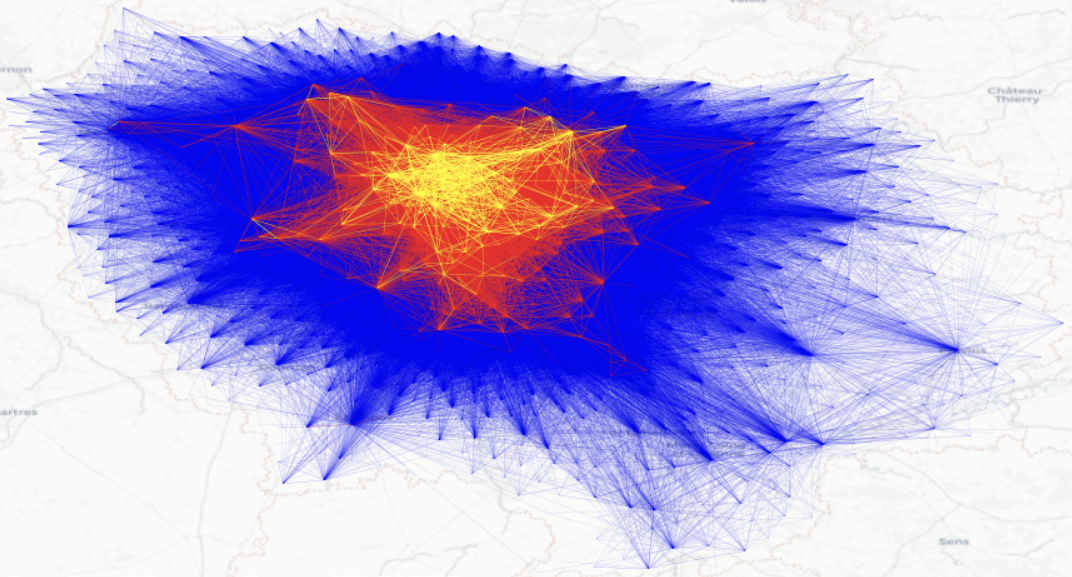}
    }
    \subfigure[Data-oriented OD flows in Paris]{
        \includegraphics[width=0.45\linewidth]{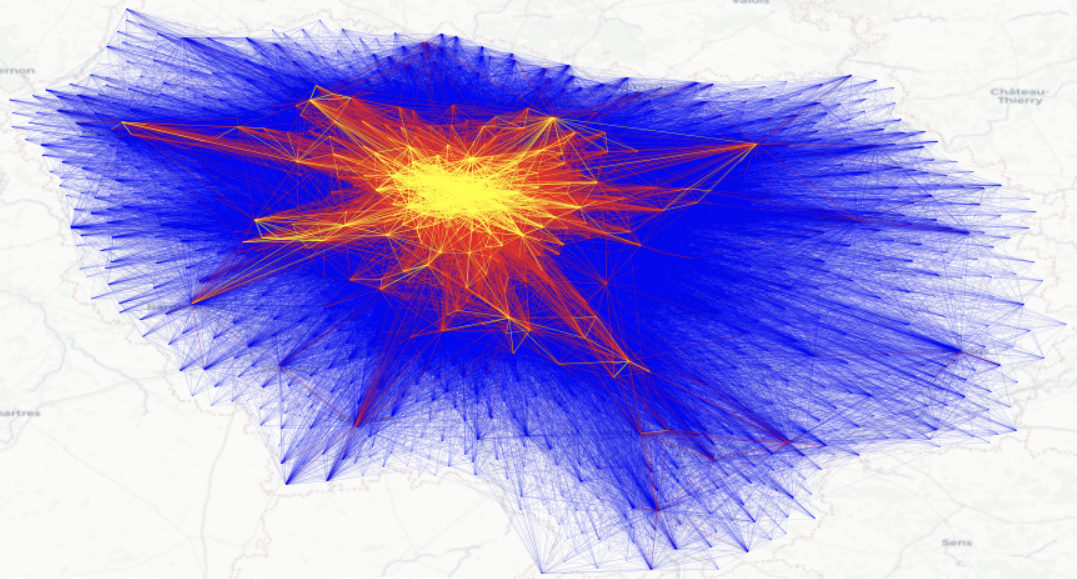}
    }
    \caption{Visualization of the generated and data-oriented OD flows in London and Paris. The generated OD flows show high consistency with the data-oriented data in both cities, demonstrating the effectiveness of GlODGen in capturing real-world mobility patterns.}
    \label{fig:spviz_eu}
\end{figure}

The Greater London and Greater Paris Metropolitan Areas are chosen as representative regions in Europe. The visually qualitative comparisons of the generation and data are shown in Figure~\ref{fig:spviz_eu}. Figures from above to below are London and Paris respectively and the left column is the generation, while the right column is OD flows from data. The generated OD flows exhibit substantial similarity to the data. The city boundary is distinctly represented by blue OD flows, while the dense red and yellow lines highlight the city center. Unlike the results observed in China, the generated flows for London successfully capture long-distance OD flows between the city center and the outskirts. In Paris, the generation accurately reflects the limited long trip flows, suggesting that our framework effectively captures Paris's centralized urban structure and efficient planning.

\subsubsection{Brazil} \label{sec:exp_latin_america}

\begin{figure}[h]
    \centering
    \subfigure[Generated OD flows in Rio de Janeiro]{
        \includegraphics[width=0.45\linewidth]{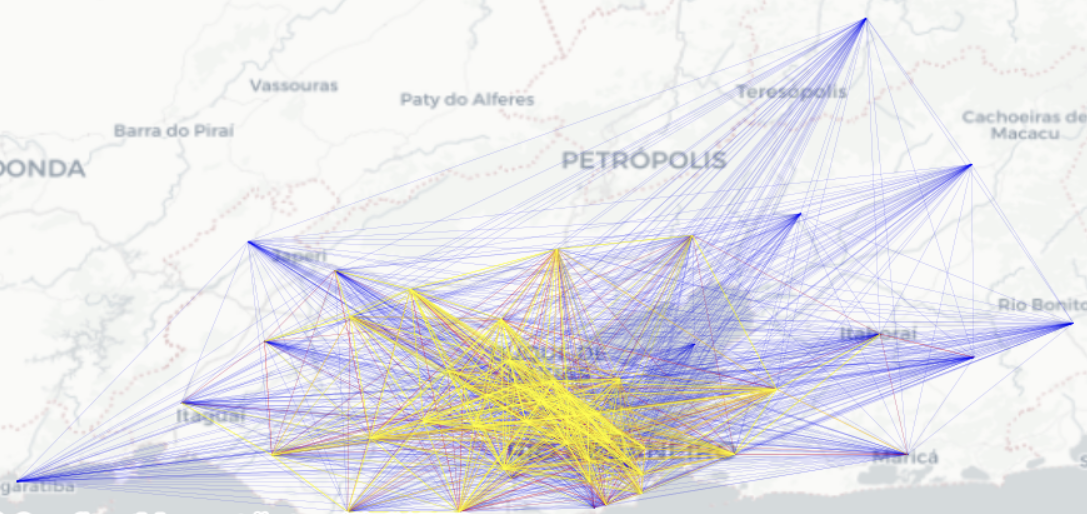}
    }
    \subfigure[Data-oriented OD flows in Rio de Janeiro]{
        \includegraphics[width=0.45\linewidth]{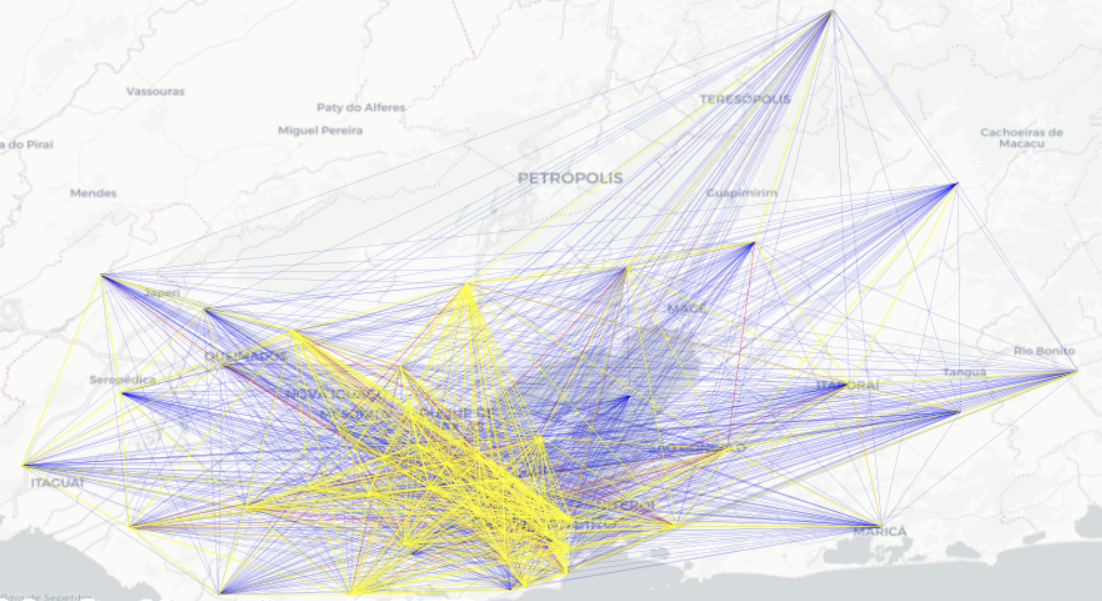}
    }
    \caption{Visualization of the generated and data-oriented OD flows in Rio de Janeiro. The generated OD flows show high consistency with the data-oriented data, demonstrating the effectiveness of GlODGen in capturing real-world mobility patterns.}
    \label{fig:spviz_br}
\end{figure}

We select the Rio de Janeiro Metropolitan Area as the representative urban area for Brazil. The spatial visualization is presented in Figure~\ref{fig:spviz_br}, demonstrating strong consistency between the generated OD flows and the data. This result highlights the similarity in urban structure and human mobility patterns between Rio de Janeiro and the United States, suggesting that global public data and OD flows from the United States may effectively support OD flow generation for Brazil.

\subsubsection{Africa} \label{sec:exp_africa}

\begin{figure}[h]
    \centering
    \subfigure[Generated OD flows in Senegal]{
        \includegraphics[width=0.45\linewidth]{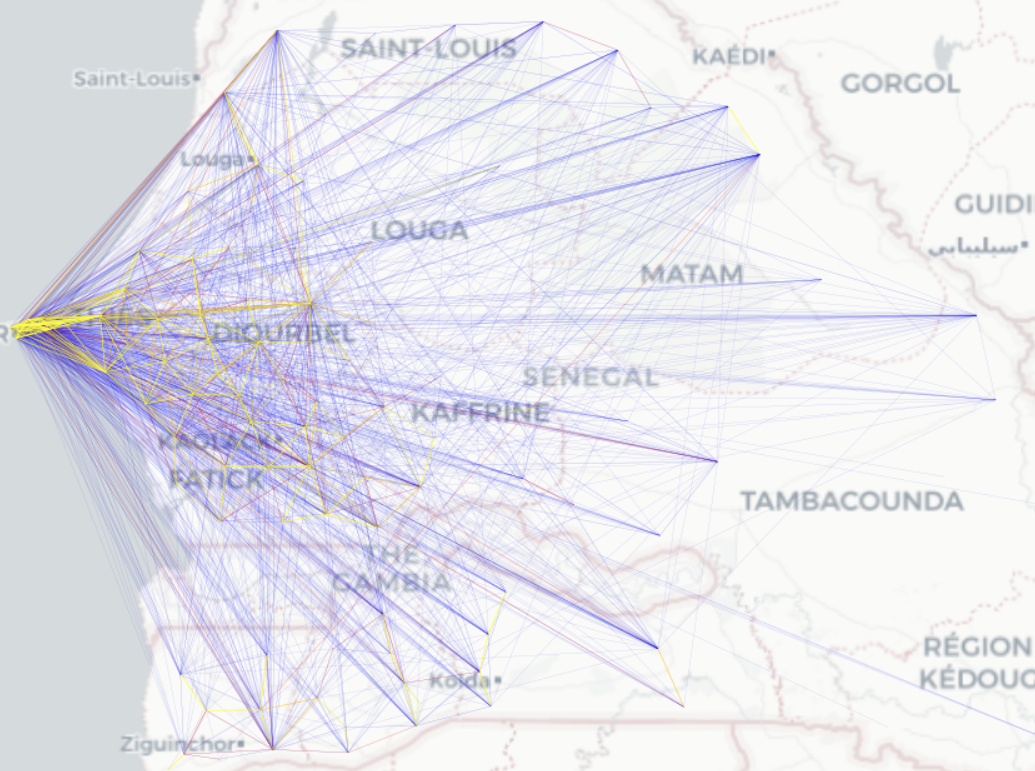}
    }
    \subfigure[Data-oriented OD flows in Senegal]{
        \includegraphics[width=0.45\linewidth]{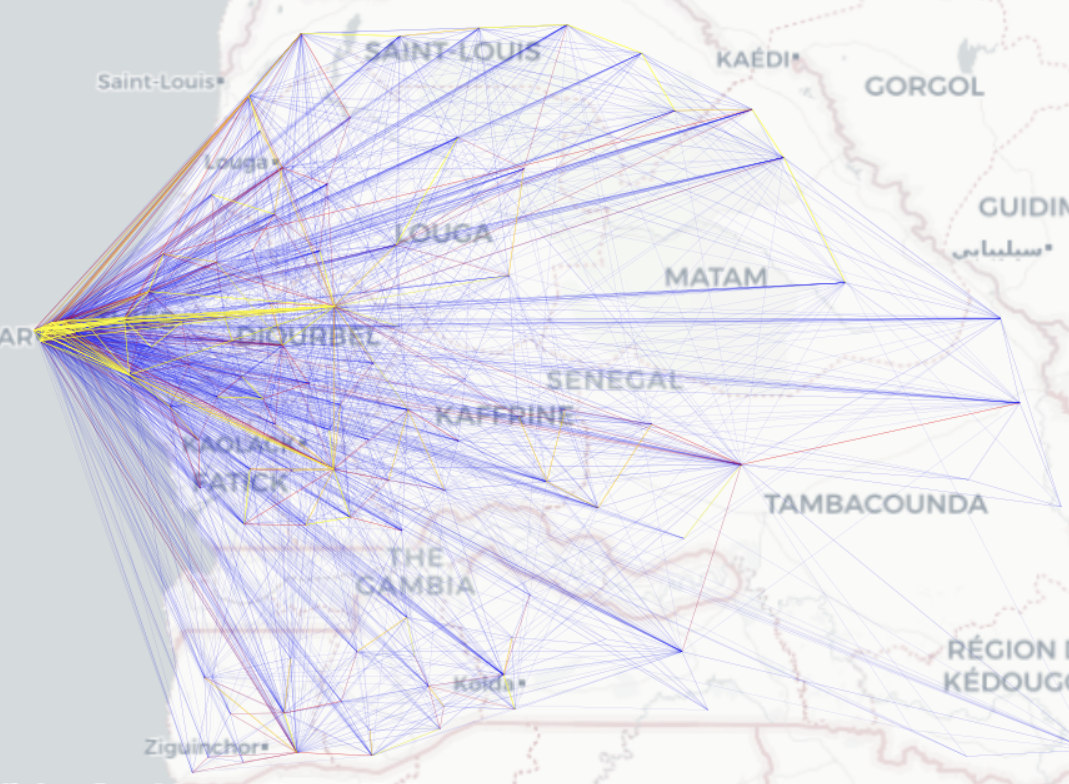}
    }
    \caption{Visualization of the generated and data-oriented OD flows in Senegal. The generated OD flows show high consistency with the data-oriented data, demonstrating the effectiveness of GlODGen in capturing real-world mobility patterns.}
    \label{fig:spviz_af}
\end{figure}

Considering the scarcity of data in Africa, we select a representative country in Africa, Senegal, to generate and evaluate our framework. OD flows of Senegal are aggregated from CDRs. The spatial visualization results are presented in Figure~\ref{fig:spviz_af}. As shown, the generated OD flows closely resemble the actual CDR data in terms of spatial distribution. Specifically, the generated OD flow data for Senegal correctly identifies the city center and the urban boundary, which highlights the capability of our framework in accurately modeling mobility patterns even in underdeveloped urban areas.

\end{document}